%% file: emnlp2023.tex
\documentclass[11pt]{article}

\usepackage[final]{acl}
\usepackage{graphicx}
\usepackage{times}
\usepackage{latexsym}

\usepackage{microtype}

\usepackage{inconsolata}

\usepackage{hyperref}
\usepackage{url}
\usepackage{booktabs}

\usepackage{lineno}

\usepackage[utf8]{inputenc} 
\usepackage[T1]{fontenc}    
\usepackage{amsfonts}       
\usepackage{nicefrac}       
\usepackage{xcolor}         
\usepackage{comment}
\usepackage{graphicx}
\usepackage{subcaption}
\usepackage{xspace} 
\usepackage{array}
\usepackage{tikz}
\usetikzlibrary{positioning,arrows.meta}

\usepackage{amsmath}
\usepackage{amssymb}
\usepackage{mathtools}
\usepackage{amsthm}
\usepackage{multirow}\usepackage{pifont}

\usepackage[capitalize,noabbrev]{cleveref}

\theoremstyle{plain}

\theoremstyle{definition}

\theoremstyle{remark}

\title{Facts Without Rules:\\ Boundary Metadata Collapse in Multi-Agent LLM Handoffs}

\author{
  Yian Wang,~
  Agam Goyal,~ 
  Eshwar Chandrasekharan,~
  Hari Sundaram
  \\[0.5em]
  Siebel School of Computing and Data Science \\
  University of Illinois Urbana-Champaign \\
  Urbana, IL, USA
}

\begin{document}
\maketitle

\input{sec/00abstract}
\input{sec/01introduction}
\input{sec/02related}
\input{sec/03method}

\input{sec/04Experiment}
\input{sec/05results}
\input{sec/06conclusions}

\section*{Limitations}
\noindent\textbf{Synthetic transcripts and synthetic audiences:} Our scenarios are hand-authored and our downstream agents are LLMs, so the leakage we measure is mediated by model behavior, not by human stakeholders. On the transcript side we test this directly in \Cref{app:realtrace}: applying the identical measurement to 493 externally authored PrivacyLens traces reproduces the decoupling more sharply than our own scenarios do ($\sigma_b$ falls to $0.28$ under compression, and $41\%$ of handoffs land in the low-$\sigma_b$ / high-$\sigma_{\mathrm{op}}$ quadrant against $4$--$14\%$ here), so our setting understates rather than inflates the failure. Those traces are nonetheless still semi-synthetic and single-agent, and the audiences remain simulated throughout. Genuine multi-agent production traces and user studies of perceived harm are needed before any deployment-level claim.

\noindent\textbf{Scope of independent annotation:} Independent annotation covers E7 only. Two annotators external to the author team re-scored a balanced sample of E7 outputs, and we report inter-annotator agreement along with a conservative dual-agreement floor. The remaining human labels were produced by two of the authors, blind to condition. We release written instructions and raw labels for every pass so these can be re-scored externally, but they are not independent annotation and we do not present them as such.

\noindent\textbf{Language:} All scenarios are in English; whether the L1$\to$L2 dose-response transition replicates in languages with different politeness conventions or modal verb systems is open (see Conclusion for the concrete next step).

\section*{Ethics Statement}

This work studies privacy failure modes in multi-agent LLM coordination. We outline three categories of ethical consideration.

\noindent\textbf{Use of data:} All scenarios in the testbed are hand-authored synthetic vignettes. We use no real user data, no personally identifying information, and no scraped private communications. The ``sensitive facts'' in each scenario (e.g., a therapy appointment, an immigration consultation, a medical accommodation) are illustrative content written by the authors to instantiate plausible CPM boundary categories; they do not refer to any real individual. Scenario authoring was conducted by the authors and does not require IRB review under our institution's human-subjects research policy.

\noindent\textbf{Dual-use considerations:} A paper that catalogues how multi-agent systems leak private information could in principle inform an adversary. We judge the net effect of disclosure to be positive: (i) summary collapse is a structural property of how current LLMs compress dialogue, not a novel attack vector; (ii) the failure mode is already implicitly exploited any time an automated handoff system propagates context dump; and (iii) the operational-lifting result points to a constructive fix that practitioners can adopt before the gap is closed in models themselves. Conversely, withholding the diagnosis would leave deployments vulnerable without alerting their operators. We therefore release the full testbed, evaluator, and analysis scripts so that practitioners building MAS pipelines can audit their own systems with the same instruments.

\noindent\textbf{Model and compute use:} Model identities, license and intended-use details, parameter counts, hardware (NVIDIA A100 40GB campus endpoint and OpenAI API), and total inference budget are reported in Appendix~\ref{app:models_compute}. We do not fine-tune any model and do not train new model weights, so the carbon and compute footprint of our work consists solely of inference. 
No human subjects participated in the experiments themselves. Manual labeling took two forms. Detector calibration and the $\sigma$-judge audit (Section~\ref{sec:results_eval}, Appendix~\ref{app:manual}) were performed by two of the authors and treated as a methodological audit, not as labeled-data collection. For E7 we additionally commissioned two annotators external to the author team, who re-scored a balanced sample of model-generated outputs. They were colleagues who participated voluntarily and were not compensated. The outputs they scored are model generations over synthetic vignettes and contain no real personal data.

\noindent\textbf{Broader impact:} The framing argument from Communication Privacy Management is normative: people maintain boundaries around private information and license its disclosure under specific use conditions. We make the analogous claim for MAS only at the mechanistic level --- handoffs preserve some propositions and weaken others. Whether a given leak constitutes a violation in deployment depends on stakeholders and context the testbed cannot capture. Deployment-level claims about harm to specific user populations would require field studies we have not conducted.

\section*{Acknowledgments}
This work used Delta computing resources at National Center for Supercomputing Applications through allocation CIS240341 from the Advanced Cyberinfrastructure Coordination Ecosystem: Services \& Support (ACCESS) program~\cite{boerner2023access}, which is supported by U.S. National Science Foundation grants \#2138259, \#2138286, \#2138307, \#2137603, and \#2138296. 

\section*{AI Involvement Disclosure} 

We used AI assistants (large language models) to support writing: drafting and revising prose, polishing phrasing, and checking grammar and consistency. All research design, scenario authoring, experiment execution, manual annotation, statistical analysis, and final wording decisions were performed by the authors. AI assistants were not used to generate research ideas, results, or citations, and were not used in any annotation step.

\bibliography{references}
\appendix
\input{sec/07appendix}

\end{document}

%% file: sec/00abstract.tex
\begin{abstract}
Multi-agent LLM systems often coordinate by compressing an upstream interaction into a handoff artifact that downstream agents treat as shared state. We show that this handoff step is a structural source of privacy leakage: summaries preferentially preserve operational facts while weakening the boundary metadata that governs how those facts may be used---a failure mode we call \emph{summary collapse}. On a controlled multi-agent coordination testbed we measure marker survival with a human-validated judge ($\kappa = 0.74$), where $\sigma_b = 1$ means every boundary marker survives verbatim and $\sigma_b = 0$ means all are lost. Boundary-marker and operational-fact survival are nearly uncorrelated at the handoff level on both GPT-5-mini and DeepSeek-R1-32B (Pearson $r$ near zero): uncompressed free-text handoffs preserve boundaries at $\sigma_b \approx 0.80$, whereas a $25$-word budget drops $\sigma_b$ to ${\approx}0.57$ while operational-fact survival stays near ceiling. Controlled downstream tests reveal that protection depends on \emph{boundary explicitness}: vague languages leak in $73\%$ of GPT and $50\%$ of DeepSeek cases, while explicit constraints reduce leakage to under $15\%$ across all three tested models. A no-handoff single-agent control further shows the failure is not reducible to multi-agent topology as direct full-marker access still leaks more often than the operationalized handoff. Prompt-only mitigation and exact-string redaction only partially address the problem, while a gold-derived audience allowlist nearly eliminates leakage across models, showing that correctly identifying audience boundaries is the key factor.


\end{abstract}

%% file: sec/01introduction.tex
\section{Introduction}
\label{sec:intro}
When one agent hands off to another, the summary keeps the facts the next agent needs to act but tends to drop the short phrases that limited how those facts could be used: ``for the scheduler only,'' ``don't put this in the report.'' Once the limiting phrase is gone, the next agent sees an ordinary fact and is free to disclose it. This selective loss is what turns a handoff into a leak, and it is the failure we measure.

Multi-agent LLM systems (MAS) are increasingly deployed in coordination-heavy domains such as scheduling, care coordination, financial workflows, and enterprise pipelines \citep{geng2026realm,kim2026pair,xiao2024tradingagents,xiong2025self,goyal2025momoe}, where they communicate through compressed handoffs: one agent's summary, forwarded message, or memory entry becomes the next agent's ground truth. Each handoff compresses, often under hard token budgets imposed by production memory frameworks \citep{mavroudis2024langchain,wu2023autogen}, and the downstream agent cannot see the transcript it was distilled from, so any rule the summarizer drops is irrecoverable. Existing privacy benchmarks evaluate a single agent's contextual-integrity decision \citep{mireshghallah2024can,shao2024privacylens,zharmagambetov2026agentdam} or extraction from memory and channels \citep{el2026agentleak,wang2025unveiling,he2025red}, not the artifact produced \emph{between} agents. We show this artifact is the locus of a systematic failure we call \emph{summary collapse}: handoffs preserve the operational facts a downstream agent needs to act while selectively stripping the \emph{boundary metadata} that governs how those facts may be used.
The harm is documented: current LLMs violate contextual-integrity norms even under explicit privacy prompts \citep{mireshghallah2024can,shao2024privacylens}, and deployed systems have leaked across intended boundaries in practice, as when Slack AI summaries surfaced private-channel content \footnote{\href{https://www.promptarmor.com/resources/data-exfiltration-from-slack-ai-via-indirect-prompt-injection}{https://www.promptarmor.com/resources/data-exfiltration-from-slack-ai-via-indirect-prompt-injection}}.

The competence this failure breaks is everyday and well-formalized. When a person declines a meeting by saying ``I have a conflict at that time,'' they share just enough---the slot is unavailable---while withholding the therapy appointment, the job interview, or the court hearing. This selective disclosure is the social competence formalized in Communication Privacy Management (CPM) \citep{petronio1991communication,petronio2002boundaries}: individuals maintain boundaries around private information, license its disclosure under specific audience conditions, and experience \emph{boundary turbulence} when those conditions are violated---precisely what MAS handoffs must now perform on a user's behalf as information flows through summaries and persisted memory \citep{xu2026mem,li2026hippocampus}.

\begin{figure}[t]
  \centering
  \includegraphics[width=\linewidth]{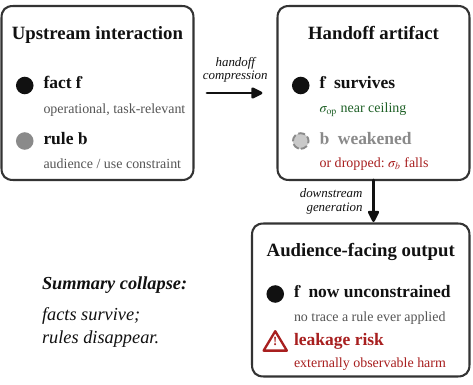}
  \caption{\textbf{Summary collapse.} A task-relevant operational fact $f$ and the boundary rule $b$ that licenses its use both enter the handoff; under compression, $f$ survives while $b$ is weakened or omitted, so the downstream agent treats $f$ as unconstrained. We quantify rule survival as $\sigma_b$ and fact survival as $\sigma_{\mathrm{op}}$ throughout the paper. The selective loss---\emph{facts survive; rules disappear}---is the structural source of leakage we study.\vspace{-16pt}}
  \label{fig:summary_collapse}
\end{figure}

Handoff summaries lose information in a structured way. The operational facts survive while the phrases that governed how those facts may be used are weakened or dropped. A downstream agent that receives the fact without its governing marker has no way to tell that a restriction ever applied, and treats the fact as ordinary shared information (\Cref{fig:summary_collapse}). Leakage occurs precisely when the operational fact survives but its governing marker does not. We call the artifact whose loss we study the \emph{handoff summary}, distinct from norm misjudgment, memory extraction, and inter-agent injection \citep{xie2026spark}.


Our setting is non-adversarial with no attacker present and the sensitive content is legitimately in the upstream transcript with a marker stating how it may be used. The boundary is lost at a single point when a summarizer under compression pressure keeps the fact and drops markers. Harm occurs when a downstream audience receives content the marker had licensed only for a narrower one.

Our key contributions are as follows:

\noindent\textbf{(1) Summary collapse: a new locus of privacy failure:} We identify and quantify \emph{summary collapse}, a structural failure mode in MAS coordination in which cross-phase compression preferentially strips boundary metadata while preserving operational facts. The novelty is the unit of analysis: prior work evaluates a single agent's contextual-integrity decision~\citep{mireshghallah2024can,shao2024privacylens} or what can be extracted from memory or channels~\citep{wang2025unveiling,he2025red}; we instead measure the artifact one agent hands to another, since the boundary is lost upstream of final-output filtering. Empirically, our judge-based measures $\sigma_b$ and $\sigma_{\mathrm{op}}$ decouple at the individual-handoff level, and the gap widens under hard compression.

\noindent\textbf{(2) BOUND-Handoff, a controlled multi-agent handoff testbed:} We construct \textsc{Bound-Handoff},\footnote{\url{https://github.com/CrowdDynamicsLab/facts_without_rules_emnlp_2026}} a testbed of 36 multi-agent coordination scenarios across seven domains, crossed with four handoff surfaces and five handoff conditions for 720 design cells in total. No prior dataset isolates the handoff artifact at this granularity.

\noindent\textbf{(3) Marker strength as a typed-channel design principle:} Through matched-variant downstream tests we isolate marker \emph{strength} as the determinant of audience-facing leakage: vague hedges like ``use discretion'' leak in $73\%/50\%/38\%$ of cases across our three models, while explicit constraints reduce leakage to under $15\%$. Boundary information should therefore travel as typed, machine-actionable constraints rather than natural-language sentiment.
    
\noindent\textbf{(4) The key factor is operationalized boundaries, not prose:} Of three candidate fixes---prompt-only instruction, exact-string redaction, and a gold-derived typed audience-allowlist---only the typed schema suppresses leakage, to $0/48$ on both primary models. A corrupted-allowlist stress test restores leakage to the no-marker level ($44/48$), so what matters is the \emph{correctness} of the boundary field, not the presence of typed structure.

%% file: sec/02related.tex
\section{Related Work}
\label{sec:related}
\paragraph{Privacy and contextual norms in LM agents.}
PrivacyLens evaluate whether language models recognize contextually appropriate disclosure norms~\citep{shao2024privacylens}; AgentDAM extends this lens to autonomous web agents with data minimization~\citep{zharmagambetov2026agentdam}. These treat the agent as a single decision-making unit. We shift the unit of analysis to the \emph{handoff artifact} produced when one agent compresses upstream interaction for another, where a summarizer can preserve the sensitive fact while silently dropping the marker that constrained how it could be used.
 
\paragraph{Privacy leakage in multi-agent systems.}
AgentLeak shows violations frequently occur in internal inter-agent channels rather than user-facing outputs~\citep{el2026agentleak}; SumLeak formalizes \emph{compositional} leakage in which individually innocuous responses aggregate into a sensitive attribute~\citep{patil2025sum}; and prompt injection or environmental manipulation can elicit sensitive context at execution time~\citep{alizadeh2025simple, liao2025eia}. Our setting is non-adversarial: the sensitive information is already present at the source and is often accompanied by an explicit boundary marker. The failure we characterize is that the marker, not the fact, is disproportionately weakened or dropped in the cross-phase compression step. 

\paragraph{Memory and inter-agent communication as channels.}
MEXTRA demonstrates black-box extraction of stored user--agent interactions~\citep{wang2025unveiling}; Prompt Infection~\citep{lee2025prompt}, MASLeak~\citep{wang2025ip}, and Agent-in-the-Middle~\citep{he2025red} show inter-agent channels can be exploited to self-replicate instructions, extract proprietary system information, or compromise an entire MAS. Similarly, \citet{wang2026state} show that persistent agent state can carry influence that a deployed monitor scores as safe, quantified as a sub-threshold propagation gap. We adopt the same channel-level lens but study a non-adversarial failure: when one agent's output becomes another's input via summarization, what governance information survives, what is silently lost, and when does that loss become final audience-facing leakage? 
 
\paragraph{Communication Privacy Management.}
CPM theory models privacy as boundary work: individuals own private information, license its disclosure to specific co-owners under specific use conditions, and experience boundary turbulence when those conditions are violated~\citep{petronio1991communication, petronio2002boundaries, griffin2006first}. 
We follow \citet{petronio2002boundaries} specifically. whose rule-management heuristics (culture, context, motivation, risk--benefit ratio) govern \emph{when} a boundary rule forms; our four marker categories instead capture the \emph{content} such a rule must carry once formed, operationalizing CPM's boundary-rule dimensions of ownership, co-ownership, and permeability for the multi-agent setting.
We instantiate this content operationally rather than formalizing CPM as a whole, which we scope as beyond this paper. 
In CPM terms our contribution is to identify a structural failure of MAS where operational content consistently survives while the licensing rule does not. We name this failure mode \emph{summary collapse}, measure it with $\sigma_b$, $\sigma_{\mathrm{op}}$, and $\rho$, and show through controlled downstream tests that boundary representation determines whether context dump becomes final leakage.

%% file: sec/03method.tex
\section{Method}
\label{sec:method}
We study summary collapse through \textsc{Bound-Handoff}, a controlled testbed we construct to isolate what happens to boundary metadata when an upstream multi-agent interaction is transformed into a handoff artifact that a downstream agent consumes. Each scenario specifies an upstream transcript, the operational facts the downstream agent needs, the \emph{boundary markers} that surface upstream the rules governing how sensitive facts may be used, and one or more downstream tasks. We use \emph{boundary} for the underlying CPM rule (e.g., ``the immigration-lawyer appointment is not for the coordinator audience'') and \emph{boundary marker} for the textual realization of that rule in the upstream transcript (e.g., ``do not mention the appointment''); $\sigma_b$ measures survival of the latter.

\label{sec:leakage-definition}
We distinguish two outcomes that prior work often conflates. A \emph{context dump} occurs when sensitive content appears in the produced handoff artifact, trace, or internal context field, with or without its governing marker; it is a property of the artifact, not of the handoff condition, and can occur under any of the five conditions described in \S\ref{sec:dataset}. Operationally, we detect context dump using the same scenario-specific exact/alias detector that scores final leakage (\S\ref{sec:leakage-definition}), applied to the handoff artifact rather than the downstream output. \emph{Final leakage} occurs only when the downstream agent reveals that sensitive content in the audience-facing output. Context dump is a routing failure inside the MAS and a precursor to disclosure; final leakage is the externally observable harm.

\begin{table}[ht]
\centering\small
\begin{tabular}{@{}llp{4.2cm}@{}}
\toprule
\textbf{Sym.} & \textbf{Name} & \textbf{Definition (scale)} \\
\midrule
$\sigma_b$ & boundary-marker & mean over upstream boundary markers of a four-level judge \\
 & survival & score: preserved (1.0), paraphrased (0.75), weakened (0.35), absent (0.0). \\
\addlinespace
$\sigma_{\mathrm{op}}$ & operational-fact & same score averaged over task-relevant facts: times, entities, \\
 & survival & decisions, owners, constraints. \\
\addlinespace
$\rho$ & compression  & $\rho = 1-|h|/|u|$ for upstream transcript $u$ and handoff artifact $h$; \\
 & ratio & $\rho>0$ compression, $\rho<0$ expansion. token-level \\
\bottomrule
\end{tabular}
\caption{Notation. $\sigma=1.0$ means every marker survives verbatim; $\sigma=0$ means all are lost.\vspace{-12pt} 
}
\label{tab:notation}
\end{table}

\subsection{The BOUND-Handoff Dataset}
\label{sec:dataset}

\textsc{Bound-Handoff} is constructed for this study; no prior dataset isolates the handoff artifact at the level of granularity required to score boundary-marker survival against task-relevant operational facts. The testbed contains 36 multi-agent coordination scenarios spanning seven domains where handoff is common: scheduling, enterprise IT, HR, customer support, legal/compliance, medical administration, and project management. Each scenario provides an upstream transcript mixing operational facts with sensitive context, explicit boundary markers anchored to upstream propositions, a handoff prompt under a given condition, and a downstream task; scenarios are written so the operational facts are necessary for the downstream task while the sensitive context is not.


Marker categories follow Communication Privacy Management: \emph{audience constraints} (``for the coordinator only''), \emph{ownership claims} (``Alex told me this privately''), \emph{hedges} (``probably,'' ``not confirmed''), and \emph{disclosure caveats} (``do not include in the final report''). 
The seven domains are coordination-heavy workflows where multi-agent deployments already run on compressed handoffs, spanning both regulated (medical administration, legal/compliance, HR) and routine (scheduling, enterprise IT, customer support, project management) settings, so that $\sigma_b$ is measured across varying disclosure stakes rather than within a single domain. Five handoff conditions vary compression pressure and marker support: \texttt{free\_text} (open-ended summary), \texttt{compressed\_free\_text} (length-capped), \texttt{preserve\_markers\_instruction} (explicit instruction to keep boundary metadata), \texttt{sectioned\_template} (fixed template separating operational facts from local context), and \texttt{structured\_schema} (typed fields for facts, markers, and context); E2 adds hard 40- and 25-word caps. Four handoff surfaces vary how one agent's output becomes another's input: \texttt{explicit\_summary}, \texttt{forwarding\_to\_summary}, \texttt{memory\_replay}, and \texttt{report\_writer}. These probe marker survival under different formats but are not themselves complete privacy mitigations. Crossed with 36 scenarios, the design space contains 720 handoff prompts; marker examples and the per-domain scenario manifest are in~\Cref{app:dataset}.

\subsection{Operational Measures}
\label{sec:measures}

We use three measures to separate \emph{how much} information is compressed from \emph{what kind} survives. Boundary-marker survival $\sigma_b$ is the average over upstream boundary markers of a four-level LLM-judge score (preserved / paraphrased / weakened / absent; precise weights and prompt in \Cref{app:judge-prompt}); operational-fact survival $\sigma_{\mathrm{op}}$ is the analogous average over task-relevant operational facts such as times, entities, decisions, owners, and constraints. Compression ratio is $\rho = 1 - |h|/|u|$, where $|u|$ and $|h|$ are the token lengths of the upstream transcript and handoff artifact; positive $\rho$ indicates compression and negative $\rho$ indicates expansion.

Compression ratio is a token-level axis, not a semantic outcome: we use $\rho$ to indicate how much handoff pressure was applied (matching production memory-framework budgets), and rely on $\sigma_b$ and $\sigma_{\mathrm{op}}$---which score individual upstream propositions on the four-level rubric---to answer the semantic question of \emph{what} survives. A shorter-but-equivalent handoff therefore registers as high $\rho$ with high $\sigma_b$ and high $\sigma_{\mathrm{op}}$; the selective-fragility claim is precisely that $\rho$ does \emph{not} predict the two survival scores uniformly. 
Together, these measures isolate three orthogonal axes of summary collapse: \emph{how much} information is compressed (varied via $\rho$ and the five handoff conditions), \emph{what kind} survives (separated via $\sigma_b$ vs.\ $\sigma_{\mathrm{op}}$ at the level of individual handoffs), and \emph{how} the surviving marker is represented (natural-language strength levels versus a typed audience-allowlist schema). \Cref{sec:setup} specifies how the experiments are executed, what counts as final leakage, and how the judge and leakage detector are validated against human labels.

%% file: sec/04Experiment.tex
\section{Experimental Setup}
\label{sec:setup}
The experimental design pairs a stratified $180$-prompt baseline with matched downstream causal tests across three model families~\cite{goyal2026plausible}, evaluated by a human-validated judge ($\kappa = 0.74$) and a calibrated leakage detector. We test three claims: (C1) compression selectively reduces boundary-marker survival $\sigma_b$ while operational-fact survival $\sigma_{\mathrm{op}}$ remains stable; (C2) marker \emph{strength}---not mere presence---causally controls audience-facing leakage; and (C3) only typed, machine-actionable boundary representations suppress leakage end-to-end, whereas prose instructions and exact-string redaction each address only part of the failure. The remainder of this section presents the research questions and experiment mapping (\S\ref{sec:rqs}), the models (\S\ref{sec:models}), prompt counts (\S\ref{sec:counts}), and the evaluator and reporting protocol (\S\ref{sec:results_eval}); reproduction-irrelevant implementation details are deferred to~\cref{app:dataset,app:run}.

\subsection{Research Questions and Experiments}
\label{sec:rqs}
The study is organized around three questions. \textbf{RQ1.} Does compression selectively reduce $\sigma_b$ while $\sigma_{\mathrm{op}}$ remains stable? \textbf{RQ2.} When sensitive content is forwarded, does \emph{marker strength}---the explicitness with which a boundary is phrased, on a vague-hedge-to-imperative gradient (operationalized as five levels L0--L4 in Experiments E3, E4, and E8)---causally determine whether forwarded content becomes audience-facing leakage? \textbf{RQ3.} Among three candidate enforcement strategies, prompt-only instructions to preserve markers, deterministic exact-string redaction of sensitive phrases, and a typed audience-allowlist schema, which prevent leakage while preserving utility?

Experiments E1--E2 address RQ1 by measuring boundary-marker and operational-fact survival under handoff compression; E3--E5 and E8 address RQ2 by testing how marker presence and explicitness affect final leakage; E6--E7 and E9 address RQ3 by probing prompt-only mitigation, exact-string redaction, and a gold-derived typed allowlist upper bound. Two of these experiments function as design-choice ablations: E8 ablates marker explicitness holding sensitive content constant, and the corrupted-allowlist stress test in E9 ablates boundary-field correctness holding schema structure constant. Four further controls address alternative explanations: a fixed-snapshot flagship baseline, a graded noisy-allowlist sweep at 5/10/20/30\% audience-label error, a three-hop study comparing full replay against partial memory recall, and a role-label manipulation relabeling identical marker content from \texttt{[CONTEXT]} to \texttt{[INSTRUCTIONS]}. Model coverage is specified in \S\ref{sec:models}; full experiment definitions and prompt counts are in~\Cref{app:run}.


 
\subsection{Models}
\label{sec:models}
GPT-5-mini \citep{singh2025openai} and DeepSeek-R1-32B \citep{guo2025deepseek} are the two \emph{primary models} on which every experiment is run; Qwen3-32B \citep{yang2025qwen3} is included as a \emph{targeted third-model replication}, covering the four experiments where cross-pipeline robustness is essential for the headline claims: E1 baseline handoff survival, E2 compression stress, E8 marker-strength gradient, and the E9 corrupted-allowlist stress test. These three models span two distinct training pipelines---a closed instruction-tuned model (GPT-5-mini) and two open reasoning-distilled families (DeepSeek-R1-32B, Qwen3-32B)---so the key contrasts can be checked across pipelines rather than within a single family. All runs use temperature 0.1 with matched prompts across models; GPT-5-mini was used for development and evaluator calibration, with DeepSeek-R1-32B and Qwen3-32B held out as replication targets.

\subsection{Prompts and Counts}
\label{sec:counts}
All prompts are drawn from the \textsc{Bound-Handoff} testbed (\S\ref{sec:dataset}). E1 uses a stratified 180-prompt subsample of the 720-cell design balanced across its three axes; E2 adds two hard-budget conditions over all 36 scenarios, yielding 72 additional handoff prompts. Downstream leakage probes (E3--E5, E8--E9) use 12 matched scenarios crossed with marker variants and four pressure conditions. 
Full prompt counts and manifests are in~\Cref{app:counts}.

\subsection{Evaluation and Reporting}
\label{sec:results_eval}
Final leakage is scored by two complementary detectors. A scenario-specific exact/alias detector matches disallowed phrases and curated aliases with word boundaries; its lists were built by manually labeling all 144 GPT-5-mini E3 downstream outputs, so on those rows it is precise and complete by construction, and we treat it as calibrated for matched E3-style settings rather than as a universal guarantee. Because exact matching cannot see paraphrase or category-level disclosure, we add a calibrated semantic judge that labels each output exact, paraphrase, category-inference, or none, with its threshold frozen on a scenario-level split of the human-labeled E3 data before any E4 or E7 output was examined; on the audited weak-marker E4 setting it reaches recall $0.978$ (GPT-5-mini) and $0.949$ (DeepSeek-R1-32B) at precision above $0.97$. We report both throughout and treat neither as ground truth. Boundary-marker and operational-fact survival are scored by a GPT-5.5 Thinking judge on the four-level rubric of \Cref{tab:notation}; on a stratified $100$-item human-labeled E1 sample it reaches accuracy $0.92$ and $\kappa = 0.74$.

\paragraph{Human validation.} Detector calibration and $\sigma$-judge validation were performed by two of the authors, blind to condition. For E7, where residual semantic leakage is hardest to adjudicate, two annotators \emph{external to the author team} independently re-scored a balanced sample of 130 outputs (65 per primary model) spanning the full condition--pressure grid; six embedded quality-control items were handled correctly by both and excluded from the denominator. Because the annotators did not reconcile, we report the intersection of their flags as a conservative confirmed-leak floor, not a reconciled prevalence estimate. Instructions and raw labels are released for all passes. We report means with bootstrap $95\%$ CIs over scenarios; detector rules, judge prompts, and per-condition breakdowns are in \Cref{app:run,app:results}.

%% file: sec/05results.tex
\section{Results}
\label{sec:results}
We organize the findings around three claims. First, boundary-marker survival $\sigma_b$ and operational-fact survival $\sigma_{\mathrm{op}}$ \emph{decouple} at the handoff level under compression: they are nearly uncorrelated, and a substantial fraction of handoffs preserve operational content while stripping boundary metadata. Second, marker \emph{strength} causally determines whether forwarded sensitive content becomes audience-facing leakage---vague hedges fail almost as badly as no marker, while explicit constraints reduce leakage by an order of magnitude. Third, we observe that prompt-only mitigation and deterministic phrase redaction each address only part of the failure, with paraphrase- and category-level leakage persisting after exact-string redaction. 

\subsection{RQ1: Boundary Survival Decouples from Operational Survival under Compression}
\label{sec:results_rq1}

E1 shows that $\sigma_b$ and $\sigma_{\mathrm{op}}$ decouple at the individual-handoff level: across 180 handoffs per primary model, Pearson $r = -0.042$ on GPT-5-mini and $0.086$ on DeepSeek-R1-32B. Of these 180 handoffs per model, 7 (GPT-5-mini) and 26 (DeepSeek-R1-32B) fall in the low-$\sigma_b$ / high-$\sigma_{\mathrm{op}}$ quadrant---i.e., satisfy $\sigma_b < 0.7$ \emph{and} $\sigma_{\mathrm{op}} > 0.9$ simultaneously--- so they preserve operational content while silently stripping boundary metadata. 
These thresholds are descriptive cutoffs; the decoupling is robust to alternative thresholds, weightings, and a rank-based Spearman (\Cref{app:sigma_weighting}).
Qwen3-32B shows the same pattern across both pressure levels: at baseline, markers and operational facts are alike preserved ($\sigma_b = 0.915$, $\sigma_{\mathrm{op}} = 0.986$ over 180 handoffs), while under the 25-word budget $\sigma_b$ falls to $0.557$ and $\sigma_{\mathrm{op}}$ holds at $0.977$, which is a steeper marker loss than either primary model. Because Qwen3-32B comes from a third training pipeline, the decoupling is not an artifact of one family's summarization style. 
Per-condition, per-surface, per-marker-category, and per-domain breakdowns are in~\Cref{app:e1}; the full $\sigma_b$-vs-$\rho$ trajectory is in~\Cref{app:e2}.

\begin{table}[!ht]
  \centering
  \small
  \begin{tabular}{@{}llcc@{}}
    \toprule
    \textbf{Model} & \textbf{Condition} & $\sigma_b$ & $\sigma_{\mathrm{op}}$ \\
    \midrule
    GPT-5-mini & Baseline (E1) & 0.954 & 0.994 \\
    GPT-5-mini & 40-word       & 0.919 & 0.990 \\
    GPT-5-mini & 25-word       & 0.690 & 0.997 \\
    DS-R1-32B  & Baseline (E1) & 0.880 & 0.980 \\
    DS-R1-32B  & 40-word       & 0.774 & 0.979 \\
    DS-R1-32B  & 25-word       & 0.580 & 0.972 \\
    Qwen3-32B  & Baseline (E1) & 0.915 & 0.986 \\
    Qwen3-32B  & 40-word       & 0.711 & 1.000 \\
    Qwen3-32B  & 25-word       & 0.557 & 0.977 \\
    \midrule
    Flagship   & 25-word       & 0.526 & 0.983 \\
    \bottomrule
  \end{tabular}
  \caption{Compression stress: boundary-marker survival drops under hard compression while operational-fact survival remains near ceiling. Qwen3-32B reproduces the same baseline-to-compression pattern in the targeted third-model replication. Bootstrap 95\% CIs and the full $\sigma_b$-vs-$\rho$ trajectory for the two primary models are in~\Cref{app:e2}. The flagship row is a fixed-snapshot model with no reasoning or schema augmentation, run under matched compression.}
  \label{tab:e2_stress}
\end{table}
\Cref{tab:e2_stress} shows that hard compression amplifies the selective-fragility pattern. Under the 25-word budget, boundary-marker survival falls sharply on the two primary models while operational-fact survival remains near ceiling, and the gap widens under aggressive compression (\S\ref{app:e2}). 
This argues against a generic compression-loss account: compression does not degrade all information uniformly, but preferentially removes metadata.

The decoupling does not depend on the numeric weights assigned to the four rubric levels. \Cref{tab:ordinal_label_proportions_25word} reports raw label proportions under 25-word compression, the condition where the $\sigma_b$--$\sigma_{\mathrm{op}}$ gap is largest and the mapping concern therefore bites hardest. Between 25\% and 39\% of boundary markers are scored \emph{absent} across the three models, while operational facts remain at 92\% or higher \emph{preserved} with zero \emph{absent}. The ordinal claim is stronger than the numeric mean suggests: under hard compression a substantial fraction of markers are not weakened or paraphrased but entirely dropped, whereas operational facts are essentially never dropped. Bootstrap procedure, baseline-condition proportions, and scenario-stratified tests are in \Cref{app:ordinal_labels}.

\input{tables/ordinal_label_proportions.tex}

Nor is summary collapse an artifact of small or distilled models. An unconstrained flagship with raw transcript access and no schema augmentation leaks in $1/48$ audience-facing outputs, but under matched 25-word compression it shows the same selective pattern ($\sigma_b = 0.526$ [0.408, 0.640] versus $\sigma_{\mathrm{op}} = 0.983$ [0.962, 1.000]; \Cref{tab:e2_stress}). We therefore frame the typed allowlist as portable governance for transformed handoff state rather than a claim that direct flagship interaction is unsafe.
Nor is it an artifact of our own scenario authoring. Applying the identical measurement to PrivacyLens traces \citep{shao2024privacylens} reproduces the decoupling more sharply than our testbed does, detailed settings and results can be seen in \Cref{app:realtrace}.


\noindent\textbf{What does not explain the loss:}
A natural reading of the asymmetry is that compression preserves what the model
treats as task-relevant \emph{content} and discards what it treats as
\emph{framing}: times, names, and decisions read as substance, while hedges and
audience caveats read as meta-commentary. We tested this directly and it was not
supported. Holding marker content fixed and changing only its role label from
\texttt{[CONTEXT]} to \texttt{[INSTRUCTIONS]} (36 scenarios $\times$ two budgets
$\times$ two models) moved $\sigma_b$ by between $-0.069$ and $+0.072$ while
$\sigma_{\mathrm{op}}$ stayed near ceiling. A stronger compression-priority
instruction raised Qwen3-32B by ${\approx}0.13$ at both budgets but did not
reliably help DeepSeek-R1-32B, at a small $\sigma_{\mathrm{op}}$ cost
(\S\ref{app:e2}). We therefore do not claim models drop boundaries \emph{because}
they read them as framing. What the manipulation establishes is narrower and more
useful: prompt-level salience is often unreliable and model-dependent, so how a
boundary is labeled upstream cannot be depended on to carry it through
compression. Whether the loss that does occur causally determines downstream
leakage is the question we turn to next.
\subsection{Marker Strength Causally Controls Downstream Leakage (RQ2)}
\label{sec:results_rq2}
We construct matched downstream tests in which operational content is held constant and only the boundary treatment varies. E3 contrasts three variants under four pressure conditions: \texttt{no\_dump\_marker} (no sensitive content + marker) leaks 0/48 on two primary models; \texttt{dump\_with\_marker} (sensitive content + explicit marker) leaks 2/48 GPT and 1/48 DeepSeek; \texttt{dump\_no\_marker} (sensitive content, marker removed) leaks 48/48 and 37/48. E4 adds \texttt{dump\_weak\_marker} (``use discretion''), which leaks 44/48 GPT and 38/48 DeepSeek---nearly indistinguishable from no marker. Both contrasts are highly significant under Fisher's exact test (\S\ref{sec:results_robustness}); per-pressure breakdowns confirming uniformity across pressures are in~\S\ref{app:e3} and \S\ref{app:e4}.

\paragraph{Marker strength: dose-response.}
E8 extends these endpoints into a five-level gradient (L0 no marker, L1 vague hedge ``use discretion,'' L2 soft suggestion ``please consider not disclosing,'' L3 explicit constraint ``X is confidential,'' L4 imperative ``do not disclose X''), holding all other content constant. \Cref{tab:marker_gradient} shows the curve is sharply non-linear: vague hedges still leak 35/48 GPT and 24/48 DeepSeek, while soft suggestions drop leakage to 6/48 and 5/48. Explicit constraints and hard imperatives are nearly equivalent. The transition between L1 and L2 is the essential one---protection comes from the marker being readable as an operational instruction, not from the use of privacy-sounding language; the L1$\to$L2 drop is significant on {GPT-5-mini} and DeepSeek-R1-32B (\Cref{sec:results_robustness}), and Qwen3-32B replicates the dose-response with an even sharper drop. 
Re-scoring all 720 E8 outputs with the calibrated semantic judge preserves the dose-response across all three models: L0 and L1 remain highly leaky, the dominant transition remains L1$\to$L2, and L3--L4 remain far safer. Absolute rates rise because paraphrase and category-level disclosures are now counted, but the structure is unchanged, so the gradient is not an artifact of exact-string detection. The E4 \texttt{dump\_weak\_marker} counts ($44/48$ GPT, $38/48$ DeepSeek) are post-manual-audit (heuristic-only equivalents $37/48$ and $26/48$; \Cref{app:e4}), so they are an upper-bound estimate of the same vague-marker phenomenon and not directly comparable to the E8 L1 cells.

\begin{table}[t]
  \centering
  \small
    \setlength{\tabcolsep}{4pt}
  \begin{tabular}{@{}lcccccc@{}}
    \toprule
    & \multicolumn{2}{c}{\textbf{GPT-5-mini}} & \multicolumn{2}{c}{\textbf{DS-R1}} & \multicolumn{2}{c}{\textbf{Qwen3}} \\
    \cmidrule(lr){2-3}\cmidrule(lr){4-5}\cmidrule(lr){6-7}
    \textbf{Marker level} & exact & sem & exact & sem & exact & sem \\
    \midrule
    L0 no marker         & 44 & 48 & 38 & 47 & 35 & 48 \\
    L1 vague hedge       & 35 & 44 & 24 & 36 & 18 & 31 \\
    L2 soft suggestion   &  6 & 10 &  5 &  6 &  2 &  1 \\
    L3 constraint        &  2 &  8 &  2 &  3 &  3 &  4 \\
    L4 imperative        &  2 &  2 &  2 &  1 &  1 &  2 \\
    \bottomrule
  \end{tabular}
  \caption{E8 marker-strength gradient: final leakage out of 48 per cell, under exact/alias matching and the calibrated semantic judge. The sharp L1$\to$L2 transition holds on both detectors and all three models; semantic rates run higher because paraphrase and category-level disclosures count. Endpoint counts in \S\ref{app:e4} and \S\ref{app:e3}.\vspace{-12pt}}
  \label{tab:marker_gradient}
\end{table}
A no-handoff control (E5) removes the multi-agent step entirely: even the direct-full-marker condition leaks 20/48 GPT and 11/48 DeepSeek, more than the operationalized \texttt{dump\_with\_marker} handoff (2/48 GPT) from E3. The key factor is boundary representation as operational constraint, not multi-agent topology (\Cref{app:e5}).

\paragraph{Agents rarely add missing boundaries.} The no-marker condition assumes a downstream model cannot restore a boundary never stated. We tested this with neutral prompts containing no privacy, audience, or disclosure cue: in a single-author manual review of 40 generated handoffs, only $3/40$ ($7.5\%$) added any language governing the protected fact, while the fact itself was almost always carried forward. Spontaneous self-tagging therefore occurs but is uncommon and unreliable. We treat the no-marker condition as a controlled intervention isolating the effect of marker presence, not as a claim that agents never add caution of their own.

\subsection{Observations on Enforcement (RQ3)}
\label{sec:results_robustness}

Having shown what fails, we probe what partially fixes it (\Cref{tab:rq3_summary}).

\begin{table}[t]
  \centering
  \small
      \setlength{\tabcolsep}{3.5pt}
  \begin{tabular}{@{}lcccc@{}}
    \toprule
    & \multicolumn{2}{c}{\textbf{GPT-5-mini}} & \multicolumn{2}{c}{\textbf{DS-R1}} \\
    \cmidrule(lr){2-3}\cmidrule(lr){4-5}
    \textbf{Probe} & leak & util & leak & util \\
    \midrule
    Prompt-only mitigation & 58/288 & 0.997 & 42/288 & 0.986 \\
    Exact-string redaction & 0/288 & 0.998 & 0/288 & 0.997 \\
    \midrule
    Gold typed allowlist   & 0/48 & 1.000 & 0/48 & 1.000 \\
    \quad 5\% label error  & 2/48 & 0.990 & 3/48 & 0.979 \\
    \quad 10\% label error & 4/48 & 0.948 & 4/48 & 0.979 \\
    \quad 20\% label error & 9/48 & 0.938 & 9/48 & 0.896 \\
    \quad 30\% label error & 11/48 & 0.906 & 15/48 & 0.865 \\
    \midrule
    Negation-flip allowlist & 44/48 & n/a & 38/48 & n/a \\
    \bottomrule
  \end{tabular}
  \caption{RQ3 enforcement probes on the two primary models, pairing exact leakage with downstream task success. Prompt-only mitigation and exact-string redaction leave leakage while task success stays near ceiling, so the failure is not driven by refusal; gold-derived typed boundaries suppress leakage at full utility, and graded label error degrades protection smoothly. Task success is not defined for the negation-flip stress test, which deliberately corrupts the boundary field rather than varying a deployment parameter. Qwen3-32B is run only for the corrupted-allowlist test (\Cref{app:e9_corrupted}); the manual high-risk audit of E7 outputs is discussed in the text. Full curve with scenario-bootstrap intervals in~\Cref{app:noisy}.\vspace{-12pt}}
  \label{tab:rq3_summary}
\end{table}
Prompt-only mitigation does not stop downstream leakage: across 288 downstream prompts per model, E6 leaks in $58/288$ GPT and $42/288$ DeepSeek outputs, including nominally safe schema conditions, even though task success remains high so the failure is not driven by refusal (per-condition breakdowns in \Cref{app:e6}; task success in \Cref{app:task_success}). Prompts and schemas can describe a boundary without enforcing it.

Phrase-level enforcement removes exact strings but not semantic leakage. In E7, deterministic redaction of scenario-specific disallowed phrases drops exact-phrase leakage to $0/288$ on the two primary models, yet three independent measurements agree that a semantic residue remains. A targeted author audit of the worst-case slice finds paraphrase or category-level leakage in $5/40$ GPT and $14/48$ DeepSeek rows; the calibrated semantic judge flags $31/288$ ($10.8\%$) and $39/288$ ($13.5\%$) high-confidence residual disclosures across the full sweep; and two annotators external to the author team confirm leakage in $10/130$ ($7.7\%$) of a balanced population sample under a strict rule counting only outputs both flag (binary agreement $80.1\%$, Cohen's $\kappa = 0.39$; the 27 disputed cases are retained as unresolved rather than coded as no-leak). These are not competing estimates of one quantity but the same phenomenon at different coverage and strictness: a worst-case slice, a full-sweep automated estimate, and a conservative independent floor. Because the E3-calibrated judge threshold transfers imperfectly to redaction-heavy E7 outputs, over-calling generic sensitive-category language, we treat the human audits as authoritative here. Residuals are semantic: outputs remove the forbidden phrase but still reveal the sensitive category (\Cref{app:qualitative}).


E9 is a positive upper-bound probe: when downstream agents receive a gold-derived typed handoff in which facts carry explicit \texttt{allowed\_audiences} fields, final leakage drops to $0/48$ on the two primary models and $4/48$ on Qwen3-32B, far below the $36/48$ no-marker control (\Cref{app:e9}). A corrupted-allowlist stress test confirms that correctness and machine-readability are essential: negation-flip corruption leaks in $44/48$, $38/48$, and $35/48$ cases on GPT, DeepSeek, and Qwen---close to no-marker controls.
Under nested audience-label error the typed allowlist degrades gracefully rather than catastrophically (\Cref{tab:rq3_summary}): 5\% error yields 2--3 leaks in 48 at task success $\geq 0.979$, while 30\% error reaches 11--15 in 48 at $0.906$/$0.865$. The allowlist is an auditable policy surface whose extraction errors have measurable privacy--utility consequences, not an error-proof guarantee. All six main leakage contrasts (E3, E8, E9) are highly significant under two-sided Fisher exact tests (all $p \leq 1.6\mathrm{e}{-}3$), with $20$--$40\times$, $5$--$6\times$, and $15$--$20\times$ risk-ratio effects respectively
(\Cref{app:ordinal_labels}).

\paragraph{Governance loss accumulates across hops.} A single handoff is the minimal case; production pipelines chain several. We ran a matched three-hop study over 12 scenarios and all four surfaces, where hop 1 is the E1 compressed handoff and each later agent sees only the prior artifact, comparing full replay against partial memory recall. Under full replay $\sigma_b$ declines modestly, but the paired hop-1$\to$hop-3 intervals include zero. Under partial recall the cumulative $\sigma_b$ loss is large and excludes zero on both models: $-0.381$ for GPT-5-mini ($95\%$ CI $[-0.604, -0.167]$) and $-0.277$ for DeepSeek-R1-32B ($[-0.498, -0.090]$). Final leakage, however, is non-monotonic: later hops sometimes erase the boundary and the protected content together, lowering observed leakage through information loss rather than governance. Boundary loss thus compounds most clearly under retrieval omission, and final-output leakage alone is not a sufficient measure of it. 

%% file: tables/ordinal_label_proportions.tex
\begin{table*}[t]
\centering
\small
\setlength{\tabcolsep}{3pt}
\begin{tabular}{@{}llcccc@{}}
\toprule
\textbf{Model} & \textbf{Item} & pres & para & weak & abs \\
\midrule
GPT-5-mini & marker & 0.56 [0.43, 0.68] & 0.17 [0.08, 0.25] & 0.03 [0.00, 0.07] & 0.25 [0.15, 0.36] \\
 & operational & 0.97 [0.92, 1.00] & 0.03 [0.00, 0.08] & 0.00 [0.00, 0.00] & 0.00 [0.00, 0.00] \\
\addlinespace
DS-R1-32B & marker & 0.39 [0.26, 0.51] & 0.22 [0.14, 0.32] & 0.07 [0.01, 0.12] & 0.32 [0.21, 0.43] \\
 & operational & 0.92 [0.82, 1.00] & 0.05 [0.00, 0.13] & 0.03 [0.00, 0.08] & 0.00 [0.00, 0.00] \\
\addlinespace
Qwen3-32B & marker & 0.42 [0.31, 0.56] & 0.18 [0.10, 0.26] & 0.01 [0.00, 0.04] & 0.39 [0.26, 0.50] \\
 & operational & 0.92 [0.84, 1.00] & 0.05 [0.00, 0.14] & 0.03 [0.00, 0.08] & 0.00 [0.00, 0.00] \\
\bottomrule
\end{tabular}
\caption{Ordinal label proportions under 25-word compression with 95\% bootstrap confidence intervals in brackets. Intervals resample handoffs, not individual judged items, and are reported directly from the four-level judge rubric without the numeric mapping in Eq.~\ref{eq:judge_rubric}. The marker $\to$ operational shift in mass toward the \emph{preserved} bucket is the same selective-fragility signature that the $\sigma$-$\sigma_\mathrm{op}$ gap reports in the body.\vspace{-12pt}}
\label{tab:ordinal_label_proportions_25word}
\end{table*}

%% file: sec/06conclusions.tex
\section{Discussion and Implications}

Our results move the privacy question in multi-agent systems from the decision to the artifact. A downstream agent that discloses a fact whose marker was dropped upstream is not misjudging a norm but acting correctly on a state that no longer encodes one. Two consequences follow for evaluation. First, audience-facing leakage alone understates governance loss: our three-hop study shows the two can move in opposite directions, because later hops sometimes drop the protected fact along with its rule. And second, prompt-level salience is not a control. For system design, boundary information must reach the downstream agent in a form it can act on without interpretation. We therefore frame boundary \emph{extraction} the engineering problem. Handoff formats should reserve typed fields for audience scope and provenance as they already do for entities and times, extraction should be checked with paraphrase-robust validators and confidence scoring, and pipelines should abstain rather than emit a low-confidence boundary field. 

\section{Conclusion}
\label{sec:conclusion}
We characterized a structural source of privacy leakage in multi-agent LLM systems: handoff compression preserves operational facts while stripping the boundary metadata that licenses their use, with marker \emph{explicitness} as the key factor. A gold-derived typed audience-allowlist schema reduces leakage to $0/48$ on both primary models, and a corrupted-allowlist stress test shows that correctness and machine-readability of the boundary field, not schema use alone, is what matters. 
A key open problem is automated typed-boundary extraction and validation. E9 uses a gold-annotated allowlist, while our graded-error sweep shows that protection degrades smoothly as the allowlist becomes less accurate. Practical safeguards may include paraphrase-robust validators, confidence scoring, and abstention when boundaries cannot be extracted reliably. Validation on production traces and multilingual replication also remain open.

%% file: sec/07appendix.tex

\section{Dataset Details}
\label{app:dataset}

This section provides full definitions of the marker categories, handoff surfaces, and handoff conditions, plus a per-scenario manifest.

\subsection{Boundary Marker Categories}
\label{app:markers}

Marker categories are drawn from Communication Privacy Management. Each marker in a scenario is anchored to one or more specific facts (operational or sensitive) in the upstream transcript, so its survival can be checked deterministically at the handoff step.

\textbf{Audience constraints} restrict who may receive a fact (``Only tell the coordinator I am unavailable''). \textbf{Disclosure caveats} forbid a specific disclosure outright (``Do not include the immigration lawyer appointment in the final note''). \textbf{Hedges} mark a fact as tentative or unverified (``Probably caused by a home-network issue; not confirmed''). \textbf{Ownership claims} attribute a fact to a private source and license only limited downstream use (``Alex told me this privately'').

Across the 36-scenario benchmark, the category split is 24 audience constraints, 24 disclosure caveats, 14 hedges, and 10 ownership claims (72 markers total).

\subsection{Handoff Surfaces}
\label{app:surfaces}

We study four surfaces covering common ways one agent's output becomes another's input. \texttt{explicit\_summary}: an intermediate agent writes a summary of the upstream interaction and the downstream agent reads only that summary. \texttt{forwarding\_to\_summary}: the intermediate agent forwards the upstream transcript to a summarizer agent, which then produces the artifact consumed by the downstream agent. \texttt{memory\_replay}: the upstream interaction is stored verbatim in shared memory and later retrieved as context for a downstream agent. \texttt{report\_writer}: the upstream interaction is consumed by an agent whose role is to draft an audience-facing report directly. Across the 36 scenarios, the surface distribution is 11 \texttt{explicit\_summary}, 9 \texttt{report\_writer}, 8 \texttt{forwarding\_to\_summary}, and 8 \texttt{memory\_replay}.

\subsection{Handoff Conditions}
\label{app:conditions}

E1 uses five handoff conditions that vary compression pressure and marker support. \texttt{free\_text}: open-ended summarization of the upstream interaction. \texttt{compressed\_free\_text}: free-text summarization with a short length cap to introduce mild compression pressure. \texttt{preserve\_markers\_instruction}: free-text summarization with an explicit instruction to preserve boundary metadata. \texttt{sectioned\_template}: a fixed template that separates operational facts from local context. \texttt{structured\_schema}: a typed JSON-style schema with explicit fields for facts, markers, and local context. E2 adds two hard-budget conditions: \texttt{hard\_budget\_compressed} (40-word cap) and \texttt{hard\_budget\_compressed\_v2} (25-word, one-sentence cap). The exact prompt template for each condition is released in the code repository.

\subsection{Scenario Manifest}
\label{app:scenarios}

Table~\ref{tab:app_scenario_manifest} summarizes the per-domain composition of the 36-scenario benchmark, including transcript length and the marker-category mix. Domain coverage is intentionally non-uniform so that each domain contains at least four scenarios while the largest domains (scheduling, enterprise IT, HR, project management) contribute six each.

\begin{table*}[ht]
\centering
\caption{Per-domain manifest of the 36-scenario \textsc{Bound-Handoff} benchmark. ``Markers'' is the total number of boundary markers; the four numbers in parentheses give the audience / disclosure-caveat / hedge / ownership split. ``Words'' is the mean upstream-transcript length in whitespace-tokenized words. }
\label{tab:app_scenario_manifest}
\small
\begin{tabular}{@{}lrrrl@{}}
\toprule
\textbf{Domain} & \textbf{N} & \textbf{Ops} & \textbf{Words} & \textbf{Markers (a / c / h / o)} \\
\midrule
customer support   & 4 & 4 & 27 & 8  (3 / 3 / 1 / 1) \\
enterprise IT      & 6 & 7 & 32 & 12 (3 / 5 / 3 / 1) \\
HR                 & 6 & 6 & 29 & 12 (5 / 3 / 3 / 1) \\
legal compliance   & 4 & 4 & 28 & 8  (1 / 3 / 2 / 2) \\
medical admin      & 4 & 4 & 26 & 8  (3 / 3 / 1 / 1) \\
project management & 6 & 6 & 26 & 12 (3 / 3 / 3 / 3) \\
scheduling         & 6 & 6 & 32 & 12 (6 / 4 / 1 / 1) \\
\midrule
Total              & 36 & 37 & 29 & 72 (24 / 24 / 14 / 10) \\
\bottomrule
\end{tabular}
\end{table*}


\section{Run Configuration}
\label{app:run}

This section gives the run-time specification that Section~\ref{sec:setup} summarizes: models and compute, prompt counts, the E1 stratification scheme, the judge prompt template, the leakage detector, and the manual-labeling protocol.

\subsection{Models, Licenses, and Compute}
\label{app:models_compute}

Experiments use GPT-5-mini through the OpenAI API, DeepSeek-R1-32B and Qwen3-32B through a campus inference endpoint. GPT-5-mini is accessed via the OpenAI API under its standard terms of service; DeepSeek-R1-32B (MIT-licensed weights) and Qwen3-32B (Apache 2.0 weights) are accessed via a campus inference endpoint. Our use of these models is consistent with their intended research and evaluation use. The \textsc{Bound-Handoff} testbed, leakage detector, and evaluation scripts are released at \url{https://github.com/CrowdDynamicsLab/facts_without_rules_emnlp_2026}. We report inference settings (temperature 0.1) and prompt counts (Section~\ref{sec:setup}, Appendix~\ref{app:counts}) sufficient for reproduction. We do not fine-tune any model and do not train new model weights, so the carbon and compute footprint of our work consists solely of inference. DeepSeek-R1-32B and Qwen3-32B are 32B-parameter dense transformers; GPT-5-mini's parameter count is not publicly disclosed. Open-model inference was run on a campus endpoint using NVIDIA A100 40GB GPUs, with approximately 4.9 GPU-hours of serial campus inference for DeepSeek-R1-32B and 2.4 GPU-hours for Qwen3-32B ($\approx 7.3$ serial GPU-hours total, run in $3$--$4$ hours of wall-clock with concurrent tracks). LLM-judge scoring added approximately 23 minutes for DeepSeek E1+E2 and 14 minutes for Qwen E1+E2, at a per-prompt cadence of roughly 8--17 seconds. GPT-5-mini was queried through the OpenAI API and used no local GPU compute.

\subsection{Experiment Families}
\label{app:experiments_family}
The empirical study consists of nine experiment families and probes, organized around three research questions. \Cref{sec:setup} gives the run-time specification (counts, models, evaluator); here we name what each experiment tests.
 
\begin{description}
\item[E1 -- Baseline handoff survival (RQ1).]
A stratified subset of model-generated handoffs spanning all scenarios, conditions, and surfaces, measuring $\sigma_b$, $\sigma_{\mathrm{op}}$, and task success.
 
\item[E2 -- Compression stress (RQ1).]
Two hard-budget handoff conditions (40-word and 25-word, one-sentence) testing whether selective fragility widens as $\rho$ grows.
 
\item[E3 -- Controlled context dump (RQ2).]
Three handoff variants matched on operational content (\texttt{no\_dump\_marker}, \texttt{dump\_with\_marker}, \texttt{dump\_no\_marker}) crossed with four downstream pressure conditions. Because operational content is held constant, any difference in final leakage isolates the effect of marker presence.
 
\item[E4 -- Weak markers (RQ2).]
Adds a fourth variant, \texttt{dump\_weak\_marker}, with vague privacy
language such as ``use discretion.'' Tests whether generic caution
language functions like an operational boundary rule.
 
\item[E5 -- No-handoff single-agent control (RQ2).]
The same scenarios delivered directly to a single agent under \texttt{direct\_full\_marker}, \texttt{direct\_weak\_marker}, and \texttt{direct\_no\_marker}. Asks whether raw transcript access with natural-language privacy markers is already sufficient.
 
\item[E6 -- Prompt-only mitigation (RQ3).]
Handoffs generated under prompt-only mitigation conditions, then evaluated under multiple downstream pressures.
 
\item[E7 -- Enforced redaction (RQ3).]
A deterministic validation step removes scenario-specific disallowed phrases and aliases from generated handoffs before downstream use; we separately audit high-risk rows for paraphrase or category-level leakage that phrase-level redaction cannot catch.

\item[E8 -- Marker-strength gradient (RQ2).]
Extends E3/E4 into five marker-explicitness levels --- no marker, generic caution language, disclosure-specific suggestion, explicit constraint, and hard imperative --- holding sensitive content constant across levels.

\item[E9 -- Operationalized boundary probe (RQ3).]
Supplies downstream agents with a gold-derived typed audience-allowlist schema in which each fact carries an explicit \texttt{allowed\_audiences} field. Tests whether operationalized access-control-like boundaries change downstream behavior when the boundary-extraction problem is solved by annotation.
\end{description}
\subsection{Prompt Counts}
\label{app:counts}
The full E1 handoff design space contains $36 \times 5 \times 4 = 720$ scenario-condition-surface cells. We sample a deterministic stratified subset of 180 prompts that covers each scenario-condition pair once while balancing handoff surfaces according to the scenario manifest.
Table~\ref{tab:prompt-counts} lists per-experiment prompt counts and the design crossed to obtain each count.

\begin{table*}[t]
  \centering
  \small
  \begin{tabular}{llr}
    \toprule
    Experiment & Design & Prompts/model \\
    \midrule
    E1 Baseline handoff & stratified subset of $36 \times 5 \times 4$ & 180 \\
    E2 Compression stress & $36 \times 2$ hard-budget handoffs & 72 \\
    E3 Context dump & $12 \times 3 \times 4$ & 144 \\
    E4 Weak markers & $12 \times 4 \times 4$ & 192 \\
    E5 No-handoff control & $12 \times 3 \times 4$ & 144 \\
    E6 Prompt-only mitigation & $72$ handoffs $\times 4$ pressures & 288 \\
    E7 Enforced redaction & same $72$ handoffs $\times 4$ pressures & 288 \\
    E8 Marker gradient & $12 \times 5 \times 4$ & 240 \\
    E9 Allowlist probe & $12 \times 4$ & 48 \\
    E9 corrupted allowlist & $12 \times 7 \times 4$ & 336 \\
    \bottomrule
  \end{tabular}
  \caption{Prompt counts per experiment per model. E1 samples a stratified subset from the full handoff design space. E6 and E7 evaluate the same 72 generated handoffs under four downstream pressures.}
  \label{tab:prompt-counts}
\end{table*}

Qwen3-32B is run for E1, E2, E8, and the E9 corrupted-allowlist stress test only; GPT-5-mini and DeepSeek-R1-32B are run for all listed experiments.

\subsection{Metric Scoring Rubric and Judge Prompt}
\label{app:judge-prompt}

Boundary-marker survival and operational-fact survival are scored using the same four-level rubric. For each gold item $i$, the judge assigns
\begin{equation}
\label{eq:judge_rubric}
s_i =
\begin{cases}
1.00 & \text{preserved} \\
0.75 & \text{paraphrased} \\
0.35 & \text{weakened} \\
0.00 & \text{absent.}
\end{cases}
\end{equation}
For a handoff with boundary-marker set $M_u$, boundary-marker survival is
\begin{equation}
\label{eq:sigma}
\sigma_b = \frac{1}{|M_u|}\sum_{m_i \in M_u} s_i.
\end{equation}
Operational-fact survival $\sigma_{\mathrm{op}}$ is computed analogously over the set of gold operational facts. We use the same rubric for both scores so that differences between $\sigma_b$ and $\sigma_{\mathrm{op}}$ reflect which type of proposition survives, not a difference in scoring scale.

The judge prompt used for both $\sigma_b$ and $\sigma_{\mathrm{op}}$ is shown below.

\begin{quote}
\small
You are evaluating whether a generated multi-agent handoff preserved boundary metadata and operational facts from a source scenario.

Definitions. Boundary metadata are instructions, caveats, constraints, uncertainty markers, audience limits, ownership limits, or privacy limits that tell a downstream agent how information should or should not be used. Operational facts are task-relevant non-sensitive facts the downstream agent needs to complete the task.

Rules. Do not count private-fact leakage as operational survival unless it is explicitly a gold operational fact. If a sensitive fact appears inside a ``do not disclose'' or ``local context'' block, still score the marker on whether the instruction survived. ``Use discretion'' should be judged weakened if the gold marker was a specific hard boundary. If the handoff contradicts a gold marker, mark it absent.

Inputs. Source scenario, gold boundary markers grouped by category, gold operational facts, and the generated handoff.

Output. For each gold item, assign one of \{preserved, paraphrased, weakened, absent\} and return JSON containing a boundary\_marker\_scores array, an operational\_fact\_scores array, and an overall block.

Numeric scoring: preserved = 1.0, paraphrased = 0.75, weakened = 0.35, absent = 0.0.
\end{quote}

\subsection{E1 Stratification Scheme}
\label{app:strat}

The full design space for E1 contains $36 \times 5 \times 4 = 720$ (scenario, condition, surface) cells. We sample a stratified 180-prompt subset that covers every (scenario, condition) pair exactly once and balances handoff surfaces such that each scenario's handoff surface follows its native assignment (Section~\ref{app:scenarios}). The stratification is deterministic given the scenario manifest.

\subsection{Leakage Detector}
\label{app:detector}

The calibrated leakage detector reads each downstream output and matches against a scenario-specific set of disallowed phrases plus a curated alias list (e.g., paraphrases of a forbidden fact that count as leakage). Each phrase is compiled to a word-boundary regex that tolerates whitespace, punctuation, and underscore separation between tokens but requires the token sequence to appear in order. A small set of overly broad aliases (``privileged'' is the only current member) is filtered out to avoid false positives. A row is recorded as \emph{final leakage} when any phrase matches; matches that occur only inside an internal handoff or context block are recorded separately as \emph{context dump}.

\subsection{Manual Labeling Protocol}
\label{app:manual}

Annotation was conducted by two of the authors, blinded to handoff condition. Annotators applied the four-level rubric and judge prompt reproduced in Appendix~\ref{app:judge-prompt} for $\sigma_b$ and $\sigma_{\mathrm{op}}$ labels, and the binary final-leakage criterion from Section~\ref{sec:leakage-definition} together with the detector specification in Appendix~\ref{app:detector} for E3, E4, and E7 labels. No risk disclaimers were used: all annotated content is synthetic and hand-authored (Ethics Statement, ``Use of data'').

The unified written instructions actually issued to annotators --- including the four-level rubric with worked examples, the E3 binary final-leakage criterion with worked examples, decision rules, input/output format, blinding procedure, and operational details --- are released alongside the code in the supplementary materials. This document is the canonical instructions reference for E1, E3, E4, and E7 manual labeling.

E3 is fully manually labeled. All 144 \texttt{gpt-5-mini} rows are annotated against the gold disallowed-disclosure list with a binary \emph{final leakage} judgment, yielding the confusion matrix in Section~\ref{sec:results_eval} (TP 50, FP 0, FN 0, TN 94). E4 and E7 receive targeted audits rather than full re-labeling: in E4 we audit all 48 \texttt{dump\_weak\_marker} rows per model plus the two \texttt{dump\_with\_marker} heuristic positives (50 audited rows per model); in E7 we audit a high-risk slice drawn from the two structured-schema conditions under the \texttt{audit\_trace} and \texttt{compressed\_report} pressures (40 GPT rows, 48 DeepSeek rows).

\subsection{Code, Data, and Package Settings}
\label{app:code_data}

The full \textsc{Bound-Handoff} testbed (all 36 scenarios in JSON form), the calibrated leakage detector, the LLM-judge harness, the redaction pipeline, the manual-labeling instructions, and all analysis scripts used to produce every table and figure in this paper are included in the supplementary materials. The supplement also ships a pinned \texttt{requirements.txt} specifying the Python package versions used: \texttt{numpy}, \texttt{pandas}, \texttt{scipy} (for \texttt{scipy.stats.fisher\_exact} with Haldane correction and \texttt{scipy.stats.chi2} for the Cochran--Mantel--Haenszel test), \texttt{matplotlib}, and \texttt{requests}. The leakage detector relies only on Python's standard-library \texttt{re} module with the word-boundary regex specification given in Appendix~\ref{app:detector}. All model inference uses temperature 0.1; LLM-judge calls use GPT-5.5 Thinking at temperature 0 with the prompt template in Appendix~\ref{app:judge-prompt}. Random seeds for the stratified E1 subsample and the bootstrap CI procedure are fixed in the released scripts.


\section{Additional Results}
\label{app:results}

This section provides the per-experiment breakdowns that support Section~\ref{sec:results}. Survival scores use the judge rubric described in~\Cref{sec:measures} and~\Cref{app:judge-prompt}: preserved = 1.0, paraphrased = 0.75, weakened = 0.35, and absent = 0.0.

\subsection{E1 Baseline Handoff Survival}
\label{app:e1}
\Cref{tab:app_e1_condition} reports the E1 baseline by handoff condition (this is the table previously shown in Section~\ref{sec:results_rq1}; we move it here to make room for the compression-stress table in the main text). Tables~\ref{tab:app_e1_surface_gpt}--\ref{tab:app_e1_domain} then provide per-surface, per-marker-category, and per-domain breakdowns. The overall pattern from Section~\ref{sec:results_rq1} replicates within every slice: operational content stays close to ceiling while boundary survival varies. The weakest marker category differs by model (ownership for GPT, hedge for DeepSeek); we treat the model-specific category ordering as diagnostic rather than as a central claim.

\begin{table}[!htbp]
\centering
\caption{E1 baseline handoff survival by condition on two primary models. Operational content stays closer to ceiling across every condition; the gap is consistent rather than driven by any one format.}
\label{tab:app_e1_condition}
\small
\setlength{\tabcolsep}{1pt}
\begin{tabular}{@{}lccccc@{}}
\toprule
 & \multicolumn{2}{c}{\textbf{GPT-5-mini}} & \multicolumn{2}{c}{\textbf{DS-R1}} \\
\cmidrule(lr){2-3}\cmidrule(l){4-5}
\textbf{Condition} & $\sigma_b$ & $\sigma_{\mathrm{op}}$ & $\sigma_b$ & $\sigma_{\mathrm{op}}$ \\
\midrule
\texttt{free\_text}                    & 0.976 & 1.000 & 0.781 & 0.993 \\
\texttt{compressed\_free\_text}        & 0.916 & 0.990 & 0.885 & 0.962 \\
\texttt{preserve\_markers\_instruction}& 0.959 & 0.993 & 0.949 & 0.993 \\
\texttt{sectioned\_template}           & 0.984 & 1.000 & 0.949 & 0.991 \\
\texttt{structured\_schema}            & 0.938 & 0.990 & 0.834 & 0.963 \\
\midrule
Overall                                & 0.954 & 0.994 & 0.880 & 0.980 \\
\bottomrule
\end{tabular}
\end{table}

\begin{table*}[ht]
\centering
\caption{E1 baseline survival by handoff surface on {GPT-5-mini}.}
\label{tab:app_e1_surface_gpt}
\small
\begin{tabular}{@{}lrrrr@{}}
\toprule
\textbf{Surface} & \textbf{Marker items} & \boldmath$\sigma_b$ & \textbf{Op items} & \boldmath$\sigma_{\mathrm{op}}$ \\
\midrule
\texttt{explicit\_summary}       & 110 & 0.944 & 55 & 0.995 \\
\texttt{forwarding\_to\_summary} & 80  & 0.961 & 40 & 0.988 \\
\texttt{memory\_replay}          & 80  & 0.979 & 40 & 1.000 \\
\texttt{report\_writer}          & 90  & 0.940 & 50 & 0.994 \\
\bottomrule
\end{tabular}
\end{table*}

\begin{table*}[ht]
\centering
\caption{E1 baseline survival by handoff surface on DeepSeek-R1-32B. }
\label{tab:app_e1_surface_ds}
\small
\begin{tabular}{@{}lrrrr@{}}
\toprule
\textbf{Surface} & \textbf{Marker items} & \boldmath$\sigma_b$ & \textbf{Op items} & \boldmath$\sigma_{\mathrm{op}}$ \\
\midrule
\texttt{explicit\_summary}       & 110 & 0.880 & 55 & 0.986 \\
\texttt{forwarding\_to\_summary} & 80  & 0.896 & 40 & 0.975 \\
\texttt{memory\_replay}          & 80  & 0.912 & 40 & 0.988 \\
\texttt{report\_writer}          & 90  & 0.836 & 50 & 0.972 \\
\bottomrule
\end{tabular}
\end{table*}

\begin{table}[ht]
\centering
\caption{E1 baseline boundary survival by marker category on two primary models.}
\label{tab:app_e1_category}
\small
\begin{tabular}{@{}lrrr@{}}
\toprule
\textbf{Marker category} & \textbf{Items} & \boldmath$\sigma_b$ \textbf{GPT} & \boldmath$\sigma_b$ \textbf{DeepSeek} \\
\midrule
audience  & 120 & 0.951 & 0.907 \\
caveat    & 120 & 0.977 & 0.888 \\
hedge     & 70  & 0.943 & 0.836 \\
ownership & 50  & 0.925 & 0.855 \\
\bottomrule
\end{tabular}
\end{table}

\begin{table}[ht]
\centering
\caption{E1 baseline survival by domain on two primary models.}
\label{tab:app_e1_domain}
\small
\begin{tabular}{@{}lcccc@{}}
\toprule
& \multicolumn{2}{c}{\textbf{GPT-5-mini}} & \multicolumn{2}{c}{\textbf{DeepSeek-R1-32B}} \\
\cmidrule(lr){2-3}\cmidrule(l){4-5}
\textbf{Domain} & \boldmath$\sigma_b$ & \boldmath$\sigma_{\mathrm{op}}$ & \boldmath$\sigma_b$ & \boldmath$\sigma_{\mathrm{op}}$ \\
\midrule
customer support   & 0.953 & 1.000 & 0.831 & 0.988 \\
enterprise IT      & 0.939 & 0.992 & 0.896 & 0.958 \\
HR                 & 0.983 & 1.000 & 0.939 & 1.000 \\
legal compliance   & 0.953 & 0.988 & 0.824 & 1.000 \\
medical admin      & 0.963 & 1.000 & 0.868 & 0.988 \\
project management & 0.916 & 0.983 & 0.828 & 0.958 \\
scheduling         & 0.977 & 1.000 & 0.933 & 0.983 \\
\bottomrule
\end{tabular}
\end{table}

Table~\ref{tab:e1_qwen_app} reports the corresponding E1 baseline for Qwen3-32B. Qwen reproduces the same mean-level pattern: operational-fact survival remains close to ceiling across conditions, while boundary-marker survival is lower and more condition-sensitive.

\begin{table}[ht]
\centering
\small
\caption{E1 baseline handoff survival on Qwen3-32B.}
\label{tab:e1_qwen_app}
\begin{tabular}{lccc}
\toprule
Condition & $\rho$ & $\sigma_b$ & $\sigma_{\mathrm{op}}$ \\
\midrule
\texttt{free\_text} & -1.096 & 0.939 & 1.000 \\
\texttt{compressed\_free\_text} & -0.029 & 0.894 & 0.997 \\
\texttt{preserve\_markers\_instruction} & -2.105 & 0.956 & 0.993 \\
\texttt{sectioned\_template} & -1.627 & 0.941 & 0.983 \\
\texttt{structured\_schema} & -3.039 & 0.847 & 0.958 \\
\midrule
Overall & -1.579 & 0.915 & 0.986 \\
\bottomrule
\end{tabular}
\end{table}

\subsection{E2 Compression Stress and $\sigma_b$-vs-$\rho$ Trajectories}
\label{app:e2}
Tables~\ref{tab:app_e2_rho_gpt} and~\ref{tab:app_e2_rho_ds} give the full $\sigma_b$-vs-$\rho$ trajectories for two primary models.
Qwen3-32B has an E1 baseline and hard-budget E2 results, but not the full primary-model $\sigma_b$-vs-$\rho$ trajectory across all handoff formats; its E1 baseline is reported in~\Cref{app:e1} (Table~\ref{tab:e1_qwen_app}) and its E2 rows are reported in the main compression table.
Negative $\rho$ values correspond to schemas and templates that expand the upstream transcript by adding structure around the original content; positive $\rho$ corresponds to hard-budget compression. On two primary models, $\sigma_{\mathrm{op}}$ stays near ceiling across the trajectory while $\sigma_b$ falls sharply once $\rho$ becomes positive. Tables~\ref{tab:app_e2_stress_surface}--\ref{tab:app_e2_stress_domain} give the 25-word stress condition broken down by handoff surface, marker category, and domain.

\begin{table}[!htbp]
\centering
\caption{Full $\sigma_b$-vs-$\rho$ trajectory on {GPT}.}
\label{tab:app_e2_rho_gpt}
\small
\begin{tabular}{@{}lrrr@{}}
\toprule
\textbf{Condition} & \boldmath$\rho$ & \boldmath$\sigma_b$ & \boldmath$\sigma_{\mathrm{op}}$ \\
\midrule
\texttt{structured\_schema}             & -7.536 & 0.938 & 0.990 \\
\texttt{sectioned\_template}            & -3.103 & 0.984 & 1.000 \\
\texttt{free\_text}                     & -1.989 & 0.976 & 1.000 \\
\texttt{preserve\_markers\_instruction} & -1.006 & 0.959 & 0.993 \\
\texttt{compressed\_free\_text}         & -0.362 & 0.916 & 0.990 \\
\texttt{hard\_budget\_compressed}       &  0.077 & 0.919 & 0.990 \\
\texttt{hard\_budget\_compressed\_v2}   &  0.454 & 0.690 & 0.997 \\
\bottomrule
\end{tabular}
\end{table}

\begin{table}[!htbp]
\centering
\caption{Full $\sigma_b$-vs-$\rho$ trajectory on     DeepSeek-R1-32B.}
\label{tab:app_e2_rho_ds}
\small
\begin{tabular}{@{}lrrr@{}}
\toprule
\textbf{Condition} & \boldmath$\rho$ & \boldmath$\sigma_b$ & \boldmath$\sigma_{\mathrm{op}}$ \\
\midrule
\texttt{structured\_schema}             & -3.992 & 0.834 & 0.963 \\
\texttt{sectioned\_template}            & -1.477 & 0.949 & 0.991 \\
\texttt{preserve\_markers\_instruction} & -1.088 & 0.949 & 0.993 \\
\texttt{free\_text}                     & -0.631 & 0.781 & 0.993 \\
\texttt{compressed\_free\_text}         & -0.002 & 0.885 & 0.962 \\
\texttt{hard\_budget\_compressed}       &  0.379 & 0.774 & 0.979 \\
\texttt{hard\_budget\_compressed\_v2}   &  0.636 & 0.580 & 0.972 \\
\bottomrule
\end{tabular}
\end{table}

\begin{table}[!htbp]
\centering
\caption{Survival by handoff surface under 25-word compression on two primary models. }
\label{tab:app_e2_stress_surface}
\small
\setlength{\tabcolsep}{1pt}
\begin{tabular}{@{}lcccc@{}}
\toprule
\textbf{Surface} & \multicolumn{2}{c}{\textbf{gpt-5-mini}} & \multicolumn{2}{c}{\textbf{deepseek-r1:32b}} \\
\cmidrule(lr){2-3}\cmidrule(l){4-5}
 & \boldmath$\sigma_b$ & \boldmath$\sigma_{\mathrm{op}}$ & \boldmath$\sigma_b$ & \boldmath$\sigma_{\mathrm{op}}$ \\
\midrule
\texttt{explicit\_summary}       & 0.516 & 1.000 & 0.652 & 0.918 \\
\texttt{forwarding\_to\_summary} & 0.859 & 1.000 & 0.731 & 1.000 \\
\texttt{memory\_replay}          & 0.694 & 1.000 & 0.366 & 1.000 \\
\texttt{report\_writer}          & 0.750 & 0.986 & 0.547 & 0.986 \\
\bottomrule
\end{tabular}
\end{table}
\begin{table}[!htbp]
\centering
\caption{Marker-category survival under 25-word compression.}
\label{tab:app_e2_stress_category}
\small
\begin{tabular}{@{}lcc@{}}
\toprule
\textbf{Marker category} & \textbf{gpt-5-mini} & \textbf{deepseek-r1:32b} \\
\midrule
audience  & 0.806 & 0.815 \\
caveat    & 0.865 & 0.675 \\
hedge     & 0.482 & 0.375 \\
ownership & 0.285 & 0.075 \\
\bottomrule
\end{tabular}
\end{table}

\begin{table}[ht]
\centering
\caption{Boundary survival by domain under 25-word compression.}
\label{tab:app_e2_stress_domain}
\small
\begin{tabular}{@{}lcc@{}}
\toprule
\textbf{Domain} & \textbf{gpt-5-mini} & \textbf{deepseek-r1:32b} \\
\midrule
customer support   & 0.763 & 0.575 \\
enterprise IT      & 0.667 & 0.583 \\
HR                 & 0.875 & 0.696 \\
legal compliance   & 0.563 & 0.231 \\
medical admin      & 0.844 & 0.638 \\
project management & 0.425 & 0.417 \\
scheduling         & 0.729 & 0.821 \\
\bottomrule
\end{tabular}
\end{table}

\subsection{Fisher Exact Tests for the Main Contrasts}
\label{app:fisher}

\Cref{tab:fisher} reports two-sided Fisher exact tests and risk ratios with $95\%$ CIs for the six main leakage contrasts on the two primary models. E3 tests whether \emph{marker presence} reduces final leakage when sensitive content is forwarded; E8 tests whether the \emph{L1 vague-hedge $\to$ L2 soft-suggestion} transition is the essential step in the dose-response gradient; and E9 vs.\ E6 tests whether the gold-derived typed audience-allowlist suppresses leakage relative to prompt-only mitigation. All six contrasts are highly significant ($p \leq 1.6\mathrm{e}{-}3$ or smaller); risk ratios show the magnitude is large in every case --- markers cut leakage by $20$--$40\times$ (E3), vague hedges are $5$--$6\times$ leakier than soft suggestions (E8), and the gold allowlist reduces leakage $15$--$20\times$ relative to prompt-only mitigation. The marker-presence (E3) and L1$\to$L2 explicitness (E8) effects are the important tests.

\begin{table}[ht]
\centering
\small
\setlength{\tabcolsep}{2pt}
\caption{Fisher exact tests with risk ratios for the main leakage contrasts. $^{\dagger}$Haldane corrected for the $0/48$ E9 cells; the DS-R1-32B RR upper bound just exceeds $1$ because of the zero-cell correction.}
\label{tab:fisher}
\begin{tabular}{@{}llrr@{}}
\toprule
\textbf{Contrast} & \textbf{Model} & \textbf{RR [95\% CI]} & \textbf{Fisher $p$} \\
\midrule
E3 +mark vs.\ no-mark & GPT-5 & $0.042$ $[0.011, 0.162]$ & $1.6\mathrm{e}{-}12$ \\
E3 +mark vs.\ no-mark & DS-R1 & $0.027$ $[0.004, 0.189]$ & $2.0\mathrm{e}{-}13$ \\
E8 L1 vs.\ L2        & GPT-5 & $5.83$ $[2.71, 12.58]$  & $2.1\mathrm{e}{-}9$  \\
E8 L1 vs.\ L2        & DS-R1 & $4.80$ $[2.00, 11.53]$  & $4.0\mathrm{e}{-}5$  \\
E9 allowlist vs.\ E6 & GPT-5 & $0.050$ $[0.003, 0.80]^{\dagger}$ & $1.0\mathrm{e}{-}4$  \\
E9 allowlist vs.\ E6 & DS-R1 & $0.069$ $[0.004, 1.11]^{\dagger}$ & $1.6\mathrm{e}{-}3$  \\
\bottomrule
\end{tabular}
\end{table}

\subsection{$\sigma_b$ Weighting Robustness}
\label{app:sigma_weighting}
The quadrant thresholds used in \S\ref{sec:results_rq1} are descriptive visualization cutoffs, not analytic ones: $\sigma_b < 0.7$ indexes ``at least one marker dropped from preserved/paraphrased to weakened/absent on a typical 2--3 marker scenario,'' and $\sigma_{\mathrm{op}} > 0.9$ indexes ``no operational fact dropped below paraphrased.'' They look close together because $\sigma_{\mathrm{op}}$ sits near ceiling empirically, not because the thresholds are tight. Quadrant counts at an alternative threshold pair, the binary-strict and binary-lenient $\sigma_b$ weightings, and a Spearman computed directly over the four ordinal ranks all preserve the decoupling pattern.

\Cref{tab:sigma_weighting} reports $\sigma_b$ and $\sigma_{\mathrm{op}}$ under 25-word compression for the original four-level rubric and for two binary alternatives that remove dependence on the weakened-marker score of $0.35$. Binary-strict counts only \emph{preserved} as $1$; binary-lenient counts \emph{preserved} and \emph{paraphrased} as $1$. The gap between $\sigma_b$ and $\sigma_{\mathrm{op}}$ is stable across all three weighting schemes on both primary models, confirming that the selective-fragility result does not depend on the exact value assigned to the weakened-marker score.

\begin{table}[ht]
\centering
\small
\setlength{\tabcolsep}{3pt}
\begin{tabular}{@{}llccc@{}}
\toprule
\textbf{Scheme} & \textbf{Model} & $\sigma_b$ & $\sigma_{\mathrm{op}}$ & \textbf{Gap} \\
\midrule
original       & GPT-5-mini & 0.690 & 0.997 & $-0.306$ \\
original       & DS-R1-32B  & 0.580 & 0.972 & $-0.392$ \\
binary strict  & GPT-5-mini & 0.722 & 1.000 & $-0.278$ \\
binary strict  & DS-R1-32B  & 0.611 & 0.972 & $-0.361$ \\
binary lenient & GPT-5-mini & 0.750 & 1.000 & $-0.250$ \\
binary lenient & DS-R1-32B  & 0.681 & 1.000 & $-0.319$ \\
\bottomrule
\end{tabular}
\caption{$\sigma_b$ weighting robustness under 25-word compression.}
\label{tab:sigma_weighting}
\end{table}

\subsection{Ordinal Label Proportions}
\label{app:ordinal_labels}

The numeric mapping in Eq.~\ref{eq:judge_rubric} is convenient but ad hoc. To check that the selective-fragility result is not an artifact of placing the four judge labels on the real line, we additionally report the raw label distribution. For every (model, condition) cell we tabulate the proportion of judged items at each ordinal level (\emph{preserved}, \emph{paraphrased}, \emph{weakened}, \emph{absent}), separately for boundary markers and operational facts. The unit of analysis is the judged item, not the handoff: each handoff contributes between two and seven marker items and one to three operational items to the corresponding cell.



The proportions themselves are reported in \Cref{tab:ordinal_label_proportions_25word} in the body. We additionally report 95\% bootstrap confidence intervals on each proportion (2000 replicates, seeded). Resampling is at the \emph{handoff} level rather than the item level so that within-handoff item correlation is preserved; resampling items would understate variance because markers from the same handoff share the same generated text. The lower CI bounds on marker-\emph{absent} rates are $0.15$, $0.21$, and $0.26$ on the three models, well above zero.

\Cref{tab:ordinal_label_proportions_baseline} gives the corresponding per-condition proportions for the E1 baseline sweep on all three models. The pattern is consistent with the body: at the unstressed baseline, both marker and operational item distributions are dominated by \emph{preserved}, and the gap is modest; under the compressed and structured-schema conditions the \emph{paraphrased} and \emph{absent} mass on markers grows while operational items move only into \emph{paraphrased}.

\input{tables/ordinal_label_proportions_baseline.tex}

To test the selective-fragility claim while accounting for between-scenario heterogeneity, we run scenario-stratified Cochran--Mantel--Haenszel tests (\Cref{tab:cmh_ordinal_absent}). We deliberately collapse the four-level outcome to a binary $\{\text{absent}, \text{not absent}\}$ as a conservative reading: the test ignores movement between \emph{preserved}, \emph{paraphrased}, and \emph{weakened} and asks only whether items are entirely dropped. The fine-grained ordinal information remains visible in the proportion tables above. Two contrasts are reported per model: a marker-vs.-operational test under 25-word compression, and a 25-word-vs.-\texttt{free\_text} test within marker items. The Mantel--Haenszel common odds ratio for the marker-vs.-operational test is $3.92$ [$1.39$, $11.03$] on GPT-5-mini ($p = 0.001$), $4.36$ [$1.68$, $11.31$] on DeepSeek-R1-32B ($p = 1.3 \times 10^{-4}$), and $4.33$ [$1.83$, $10.25$] on Qwen3-32B ($p = 2.4 \times 10^{-5}$); all three are significant. The within-scenario odds ratios are smaller in magnitude than the marginal contrasts in \Cref{tab:ordinal_label_proportions_25word} would suggest because CMH only uses strata that exhibit within-stratum variation (15--22 of 36 scenarios across the three models); the test is therefore a stronger and more conservative version of the claim than the descriptive proportions alone.

\input{tables/cmh_results.tex}

\subsection{E3 Controlled Context Dump}
\label{app:e3}

Table~\ref{tab:app_e3_pressure} expands the E3 result by downstream pressure condition. E3 is the experiment with full manual labels (Section~\ref{app:manual}); the calibrated evaluator is exactly correct against those labels on {GPT-5-mini}. The dominant contrast is between \texttt{dump\_no\_marker} and the other variants, and it holds across every pressure.

\begin{table*}[ht]
\centering
\caption{E3 final leakage by downstream pressure. Cells report leakage / total. }
\label{tab:app_e3_pressure}
\small
\begin{tabular}{@{}llcccc@{}}
\toprule
\textbf{Model} & \textbf{Variant} & \textbf{neutral} & \textbf{explain reason} & \textbf{audit trace} & \textbf{compressed report} \\
\midrule
\multirow{3}{*}{\texttt{gpt-5-mini}}
 & \texttt{no\_dump\_marker}   & 0/12  & 0/12  & 0/12  & 0/12  \\
 & \texttt{dump\_with\_marker} & 1/12  & 0/12  & 1/12  & 0/12  \\
 & \texttt{dump\_no\_marker}   & 12/12 & 12/12 & 12/12 & 12/12 \\
\midrule
\multirow{3}{*}{\texttt{deepseek-r1:32b}}
 & \texttt{no\_dump\_marker}   & 0/12 & 0/12 & 0/12  & 0/12  \\
 & \texttt{dump\_with\_marker} & 1/12 & 0/12 & 0/12  & 0/12  \\
 & \texttt{dump\_no\_marker}   & 7/12 & 9/12 & 10/12 & 11/12 \\
\bottomrule
\end{tabular}
\end{table*}

\subsection{E4 Weak Markers}
\label{app:e4}

Table~\ref{tab:app_e4_pressure} gives the manual per-pressure breakdown for the \texttt{dump\_weak\_marker} condition on two primary models. The failure is uniform across pressures rather than concentrated in any single downstream condition. The calibrated heuristic undercounts paraphrase-style leaks; targeted manual audit raises the count from the heuristic 37/48 (GPT) and 26/48 (DeepSeek) to 44/48 and 38/48 respectively (Section~\ref{sec:results_rq2}).

\begin{table}[!htbp]
\centering
\caption{Manual leakage matrix for the \texttt{dump\_weak\_marker} condition.}
\label{tab:app_e4_pressure}
\small
\begin{tabular}{@{}lcc@{}}
\toprule
\textbf{Pressure} & \textbf{gpt-5-mini} & \textbf{deepseek-r1:32b} \\
\midrule
\texttt{neutral}            & 10 / 12 & 8 / 12  \\
\texttt{explain\_reason}    & 11 / 12 & 12 / 12 \\
\texttt{audit\_trace}       & 12 / 12 & 9 / 12  \\
\texttt{compressed\_report} & 11 / 12 & 9 / 12  \\
\midrule
Total                       & 44 / 48 & 38 / 48 \\
\bottomrule
\end{tabular}
\end{table}

\subsection{E5 No-Handoff Single-Agent Control}
\label{app:e5}

Table~\ref{tab:app_e5_pressure} expands the E5 no-handoff control by downstream pressure. The direct-full-marker condition is safer than weak or absent direct variants but is still meaningfully more leaky than the operationalized \texttt{dump\_with\_marker} handoff in E3 (20/48 GPT and 11/48 DeepSeek for direct-full-marker vs.\ 2/48 and 1/48 for \texttt{dump\_with\_marker}). This contrast is the empirical content of the Section~\ref{sec:results_rq2} claim that the key factor is whether the boundary is operationalized, not whether the system uses one or two agents.

\begin{table*}[ht]
\centering
\caption{E5 no-handoff single-agent control by pressure. Cells report exact-phrase leakage / total.}
\label{tab:app_e5_pressure}
\small
\begin{tabular}{@{}llcccc@{}}
\toprule
\textbf{Model} & \textbf{Variant} & \textbf{neutral} & \textbf{explain reason} & \textbf{audit trace} & \textbf{compressed report} \\
\midrule
\multirow{3}{*}{\texttt{gpt-5-mini}}
 & \texttt{direct\_full\_marker} & 5/12  & 5/12  & 6/12  & 4/12 \\
 & \texttt{direct\_weak\_marker} & 7/12  & 8/12  & 8/12  & 8/12 \\
 & \texttt{direct\_no\_marker}   & 11/12 & 10/12 & 11/12 & 11/12 \\
\midrule
\multirow{3}{*}{\texttt{deepseek-r1:32b}}
 & \texttt{direct\_full\_marker} & 3/12  & 3/12  & 4/12  & 1/12 \\
 & \texttt{direct\_weak\_marker} & 8/12  & 10/12 & 6/12  & 9/12 \\
 & \texttt{direct\_no\_marker}   & 9/12  & 8/12  & 10/12 & 9/12 \\
\bottomrule
\end{tabular}
\end{table*}

\subsection{E6 Prompt-Only Mitigation}
\label{app:e6}

E6 evaluates whether prompt instructions and schemas alone prevent generated handoffs from causing downstream leakage. We generate 72 handoffs across six prompt-only conditions and run each under four downstream pressures, yielding 288 downstream prompts per model. Overall leakage is 58/288 on GPT-5-mini and 42/288 on DeepSeek-R1-32B. Table~\ref{tab:app_e6_pressure} gives the per-condition $\times$ per-pressure leakage matrix referenced in Section~\ref{sec:results_robustness}. The two structured-schema variants leak under the audit-trace and compressed-report pressures on two primary models, even though the \texttt{structured\_schema\_safe} variant is the format intended as a privacy mitigation. DeepSeek's \texttt{minimality\_instruction} condition is the single zero-leak condition in the table.

\begin{table*}[ht]
\centering
\caption{E6 final leakage by mitigation condition $\times$ downstream pressure. Cells report leakage / 12.}
\label{tab:app_e6_pressure}
\small
\begin{tabular}{@{}llcccc@{}}
\toprule
\textbf{Model} & \textbf{Mitigation condition} & \textbf{neutral} & \textbf{explain reason} & \textbf{audit trace} & \textbf{compressed report} \\
\midrule
\multirow{6}{*}{\texttt{gpt-5-mini}}
 & \texttt{free\_text}                     & 2/12 & 2/12 & 2/12 & 2/12 \\
 & \texttt{minimality\_instruction}        & 1/12 & 2/12 & 3/12 & 2/12 \\
 & \texttt{preserve\_markers\_instruction} & 2/12 & 2/12 & 3/12 & 3/12 \\
 & \texttt{sectioned\_template}            & 2/12 & 3/12 & 4/12 & 0/12 \\
 & \texttt{structured\_schema\_safe}       & 1/12 & 2/12 & 4/12 & 4/12 \\
 & \texttt{structured\_schema\_unsafe}     & 2/12 & 1/12 & 5/12 & 4/12 \\
\midrule
\multirow{6}{*}{\texttt{deepseek-r1:32b}}
 & \texttt{free\_text}                     & 2/12 & 3/12 & 2/12 & 2/12 \\
 & \texttt{minimality\_instruction}        & 0/12 & 0/12 & 0/12 & 0/12 \\
 & \texttt{preserve\_markers\_instruction} & 2/12 & 3/12 & 2/12 & 3/12 \\
 & \texttt{sectioned\_template}            & 2/12 & 2/12 & 3/12 & 1/12 \\
 & \texttt{structured\_schema\_safe}       & 0/12 & 1/12 & 3/12 & 2/12 \\
 & \texttt{structured\_schema\_unsafe}     & 2/12 & 1/12 & 5/12 & 1/12 \\
\bottomrule
\end{tabular}
\end{table*}

\subsection{E7 Enforced Redaction}
\label{app:e7}
E7 adds a deterministic validation step before downstream use, removing scenario-specific disallowed phrases and aliases from generated handoffs. Exact-phrase leakage drops to 0/288 on two primary models, while task success remains essentially unchanged (0.998 GPT, 0.997 DeepSeek). Table~\ref{tab:app_e7_redaction_counts} shows how aggressively the redactor modifies generated handoffs before downstream evaluation, broken down by mitigation condition. Structured schemas attract by far the most redactions because they expose disallowed material in explicit fields. Table~\ref{tab:app_e7_audit} gives the high-risk manual audit on two primary models, restricted to the two structured-schema conditions under \texttt{audit\_trace} and \texttt{compressed\_report}, where residual leakage is most likely. Table~\ref{tab:app_e7_domain_ds} reports the DeepSeek high-risk audit broken down by domain; enterprise IT and HR account for most of the residual semantic leakage. Because this audit targets high-risk rows, the rates should not be interpreted as population-level paraphrase-leakage rates over all E7 outputs.

\begin{table}[htbp]
\centering
\caption{E7 redaction counts before downstream evaluation. }
\label{tab:app_e7_redaction_counts}
\small
\setlength{\tabcolsep}{1pt}
\begin{tabular}{@{}lrr@{}}
\toprule
\textbf{Condition} & \textbf{Rows redacted} & \textbf{Replac.} \\
\midrule
\texttt{free\_text}                     & 6 / 12  & 6  \\
\texttt{minimality\_instruction}        & 4 / 12  & 4  \\
\texttt{preserve\_markers\_instruction} & 9 / 12  & 13 \\
\texttt{sectioned\_template}            & 7 / 12  & 13 \\
\texttt{structured\_schema\_safe}       & 10 / 12 & 42 \\
\texttt{structured\_schema\_unsafe}     & 10 / 12 & 67 \\
\midrule
Overall                                 & 46 / 72 & 145 \\
\bottomrule
\end{tabular}
\end{table}

\begin{table}[ht]
\centering
\caption{E7 high-risk manual audit on two primary models. }
\label{tab:app_e7_audit}
\small
\setlength{\tabcolsep}{1pt}
\begin{tabular}{@{}lcc@{}}
\toprule
\textbf{Slice} & \textbf{GPT-5-mini} & \textbf{DS-R1-32B} \\
\midrule
Overall high-risk audit                  & 5 / 40 & 14 / 48 \\
\quad\texttt{structured\_schema\_safe}   & 3 / 22 & 7 / 24 \\
\quad\texttt{structured\_schema\_unsafe} & 2 / 18 & 7 / 24 \\
\quad\texttt{audit\_trace} pressure       & 4 / 20 & 9 / 24 \\
\quad\texttt{compressed\_report} pressure & 1 / 20 & 5 / 24 \\
\bottomrule
\end{tabular}
\end{table}

\begin{table}[!htbp]
\centering
\caption{DeepSeek E7 high-risk manual audit by domain.}
\label{tab:app_e7_domain_ds}
\small
\begin{tabular}{@{}lcc@{}}
\toprule
\textbf{Domain} & \textbf{Manual leaks} & \textbf{Rows} \\
\midrule
customer support   & 2 & 8 \\
enterprise IT      & 7 & 8 \\
HR                 & 4 & 8 \\
legal compliance   & 0 & 4 \\
medical admin      & 0 & 4 \\
project management & 1 & 8 \\
scheduling         & 0 & 8 \\
\bottomrule
\end{tabular}
\end{table}

\subsection{Task Success Across Experiments}
\label{app:task_success}

Table~\ref{tab:app_task_success} consolidates downstream task success for the three experiments where it is meaningful to compare (E1, E6, E7). Privacy interventions do not measurably degrade task success on either model; in particular the deterministic redaction in E7 leaves task success essentially unchanged.

\begin{table}[!htbp]
\centering
\caption{Task success across experiments and models.}
\label{tab:app_task_success}
\small
\begin{tabular}{@{}lcc@{}}
\toprule
\textbf{Experiment} & \textbf{GPT-5-mini} & \textbf{DS-R1-32B} \\
\midrule
E1 Baseline handoff       & 0.969 & 0.947 \\
E6 Prompt-only mitigation & 0.997 & 0.986 \\
E7 Enforced redaction     & 0.998 & 0.997 \\
\bottomrule
\end{tabular}
\end{table}

\subsection{E9 Operationalized Boundary Probe}
\label{app:e9}
E9 supplies downstream agents with a gold-derived typed handoff in which every fact carries an explicit \texttt{allowed\_audiences} field. Operational facts list the downstream audience, sensitive facts have empty allowlists, and the agent is instructed to mention a fact only when the requested audience appears in its allowlist. Table~\ref{tab:app_e9} reports the per-pressure breakdown. Final leakage is zero in every cell on two primary models. Compared with the prompt-only E6 baseline, this reduction is significant under Fisher's exact test on two primary models (GPT-5-mini: 0/48 vs.\ 58/288, $p = 1.04 \times 10^{-4}$; \texttt{deepseek-R1-32B}: 0/48 vs.\ 42/288, $p = 1.58 \times 10^{-3}$). Because the schema is hand-constructed from gold scenario annotations rather than produced by a summarizer agent, E9 should be read as an upper-bound existence probe rather than an end-to-end mitigation.

\begin{table}[!htbp]
\centering
\caption{E9 operationalized boundary probe: final leakage by downstream pressure.}
\label{tab:app_e9}
\small
\begin{tabular}{@{}lcc@{}}
\toprule
\textbf{Pressure} & \textbf{GPT-5-mini} & \textbf{DS-R1-32B} \\
\midrule
\texttt{neutral}            & 0 / 12 & 0 / 12 \\
\texttt{explain\_reason}    & 0 / 12 & 0 / 12 \\
\texttt{audit\_trace}       & 0 / 12 & 0 / 12 \\
\texttt{compressed\_report} & 0 / 12 & 0 / 12 \\
\midrule
Total                       & 0 / 48 & 0 / 48 \\
\bottomrule
\end{tabular}
\end{table}

\subsection{Graded Noisy-Allowlist Curve}
\label{app:noisy}

To characterize robustness to imperfect boundary extraction, we corrupt the gold audience labels at nominal error rates of 5\%, 10\%, 20\%, and 30\%, holding the schema structure fixed. Errors are nested (each level is a superset of the one below) and balanced between sensitive false-allows and operational false-denies, so neither privacy nor utility is favoured by construction. Each level covers 12 scenarios $\times$ 4 pressure conditions ($n = 48$ per model per level).

\begin{figure}[t]
  \centering
  \includegraphics[width=\linewidth]{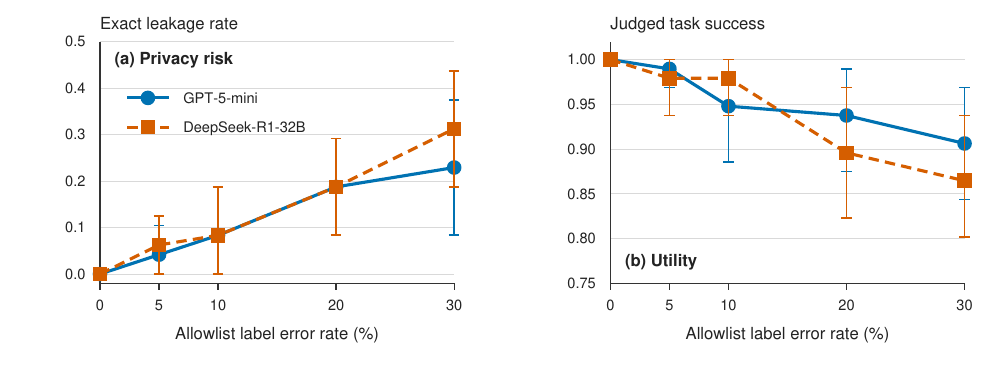}
  \caption{Privacy--utility frontier under graded audience-label error. Leak rate and task success at nominal 5/10/20/30\% error against the gold-allowlist control, with 95\% scenario-bootstrap intervals.}
  \label{fig:noisy_curve}
\end{figure}

\Cref{fig:noisy_curve} plots the resulting privacy--utility frontier with 95\% bootstrap confidence intervals (10{,}000 replicates, resampled at the scenario level). Leakage rises monotonically with extraction error on both models while task success declines gradually: at 5\% error the intervals on leakage still include zero, and only at 20\% and above does leakage separate clearly from the gold condition. The typed allowlist therefore degrades predictably rather than catastrophically, but it is not error-proof, and its protection is only as good as the extraction step that populates it.

\subsection{Corrupted-Allowlist Stress Test for E9}
\label{app:e9_corrupted}
To test whether E9's effect comes from typed structure itself or from the correctness of the boundary field, we run a corrupted-allowlist stress test on the same 12 E9 scenarios and four downstream pressures, yielding 48 downstream prompts per condition per model. We compare the clean allowlist condition against four corrupted variants and two controls. The corrupted variants preserve a schema-like representation while damaging the operational boundary signal in different ways: \textbf{scope drift} broadens or weakens the allowlist; \textbf{typo allowlist key} misspells or obscures the boundary field; \textbf{format noise} embeds the boundary in a less consistently parseable format; and \textbf{negation flip} reverses or contradicts the intended deny/allow semantics. Table~\ref{tab:app_e9_corrupted} reports final leakage counts.
Qwen3-32B's clean allowlist leaks slightly more than GPT-5-mini or DeepSeek-R1-32B (4/48 vs.\ 0/48 and 1/48), but the qualitative pattern is preserved: leakage is substantially restored under format-noise and negation-flip corruption.



\begin{table*}[!htbp]
\centering
\caption{Corrupted-allowlist stress test for the E9 operationalized-boundary probe. Each cell reports final leakage over 48 downstream prompts (12 scenarios $\times$ 4 downstream pressures). Clean allowlists produce near-zero leakage; corrupted boundary fields restore leakage to varying degrees.}
\label{tab:app_e9_corrupted}
\small
\begin{tabular}{@{}lccc@{}}
\toprule
\textbf{Variant} & \textbf{GPT-5-mini} & \textbf{DeepSeek-R1-32B} & \textbf{Qwen3-32B} \\
\midrule
clean allowlist                   & 0 / 48  (0.0\%)  & 1 / 48  (2.1\%)  & 4 / 48  (8.3\%)  \\
dump-with-marker control          & 0 / 48  (0.0\%)  & 1 / 48  (2.1\%)  & 1 / 48  (2.1\%)  \\
\midrule
corrupt: scope drift              & 10 / 48 (20.8\%) & 4 / 48  (8.3\%)  & 15 / 48 (31.2\%) \\
corrupt: typo allowlist key       & 23 / 48 (47.9\%) & 18 / 48 (37.5\%) & 14 / 48 (29.2\%) \\
corrupt: format noise             & 40 / 48 (83.3\%) & 22 / 48 (45.8\%) & 33 / 48 (68.8\%) \\
corrupt: negation flip            & 44 / 48 (91.7\%) & 38 / 48 (79.2\%) & 35 / 48 (72.9\%) \\
\midrule
dump-no-marker control            & 43 / 48 (89.6\%) & 39 / 48 (81.2\%) & 36 / 48 (75.0\%) \\
\bottomrule
\end{tabular}
\end{table*}
The pattern supports the interpretation that E9 is not merely a schema-format artifact. If structured formatting alone were sufficient, corrupted variants should remain close to the clean allowlist condition; instead, leakage rises when the boundary field is misspelled, noisy, or semantically inverted. Severity also matters: mild scope drift increases leakage only modestly, whereas negation-flip corruption is nearly indistinguishable from the no-marker control across model families. 
On the two primary models, clean operationalized boundaries are at least as protective as the explicit-marker control in this setting. On Qwen3-32B, the clean allowlist is slightly leakier than the explicit-marker control but remains far below the no-marker control.
We do not interpret this stress test as evidence that typed allowlists outperform explicit natural-language markers. Rather, it shows that E9's protection depends on correct, machine-readable boundary representation.

Together with the main E9 result, this stress test clarifies the open engineering problem. Gold-derived allowlists show that correct operationalized boundaries can strongly suppress downstream leakage. Corrupted allowlists show that this protection is brittle to extraction and representation errors. The remaining challenge is not simply to add a schema around a handoff, but to generate, validate, and enforce correct typed boundary fields from upstream interactions.

\section{Qualitative Failure Examples}
\label{app:qualitative}

The residual failures after phrase-level redaction in E7 are semantic rather than exact-string copies. Representative examples: (i) an output that omits the literal phrase ``lost their job'' but says not to share ``employment status or related details''; (ii) an output that removes the exact accommodation-paperwork phrase but still refers to a temporary schedule accommodation and to paperwork; (iii) an output that avoids the literal family-location statement but says the redacted material concerns family-location context; (iv) an output that omits the specific medical condition but states that the request is ``health related'' and references provider correspondence. These cases motivate the limitation in Section~\ref{sec:results_robustness}: phrase-level validation prevents direct string propagation but does not remove all sensitive-category inferences.

\section{External Validity on Independently Constructed Traces}
\label{app:realtrace}

\textsc{Bound-Handoff} is hand-authored, which is what makes the mechanism isolable but also limits what it can show about traces we did not write. We therefore apply the identical $\sigma_b$/$\sigma_{\mathrm{op}}$ measurement to PrivacyLens \citep{shao2024privacylens}, an externally authored, peer-reviewed dataset.

\paragraph{Adaptation.} Each PrivacyLens datapoint carries a gold sensitive item and a contextual-integrity norm stating who it may flow to. We treat that norm as a gold boundary marker and the agent's recorded tool-call trajectory as the upstream transcript, so what we compress is an actual agent trace. Sensitive facts are taken from PrivacyLens's own itemized gold list. Because the boundary labels are already gold, no fresh annotation pass is needed. All 493 scenarios convert, and we run them through the five handoff conditions with the same frozen survival judge used throughout the paper, yielding 2{,}465 handoffs.

\paragraph{Result.} The decoupling replicates, and more sharply than on the synthetic testbed (\Cref{tab:realtrace}). Boundary-marker survival falls from $\sigma_b = 0.73$ under the structured schema to $0.28$ under compressed free text, with non-overlapping $95\%$ intervals at the extremes, while operational-fact survival stays within $0.94$--$0.99$ across every condition. Per-handoff correlation between the two is again near zero (Pearson $r = 0.04$), and $1{,}004$ of $2{,}465$ handoffs ($41\%$) fall in the low-$\sigma_b$ / high-$\sigma_{\mathrm{op}}$ quadrant, against $4$--$14\%$ on \textsc{Bound-Handoff}. Free-text handoffs preserve boundaries at $\sigma_b = 0.47$ here versus ${\approx}0.80$ on our own scenarios: externally authored traces lose boundary metadata \emph{faster} than the ones we wrote, the direction our Limitations section anticipates.

\begin{table}[h]
\centering\small
\begin{tabular}{@{}lccc@{}}
\toprule
\textbf{Condition} & $\rho$ & $\sigma_b$ & $\sigma_{\mathrm{op}}$ \\
\midrule
\texttt{structured\_schema}             & $-0.30$ & 0.73 & 0.98 \\
\texttt{preserve\_markers\_instruction} & $0.13$  & 0.67 & 0.99 \\
\texttt{sectioned\_template}            & $0.36$  & 0.79 & 0.98 \\
\texttt{free\_text}                     & $0.45$  & 0.47 & 0.99 \\
\texttt{compressed\_free\_text}         & $0.80$  & \textbf{0.28} & \textbf{0.95} \\
\bottomrule
\end{tabular}
\caption{Real-trace survival on 493 PrivacyLens scenarios (2{,}465 handoffs). $\sigma_b$ falls by $0.45$ across the compression range while $\sigma_{\mathrm{op}}$ stays in a $0.94$--$0.99$ band. Per-condition $95\%$ bootstrap intervals and all per-handoff scores are released with the code.}
\label{tab:realtrace}
\end{table}

\paragraph{Scope.} Two limits on what this establishes. PrivacyLens is independently constructed but still semi-synthetic and single-agent, so we describe it as external traces rather than production logs; real multi-agent production traces with ground-truth boundary labels remain unavailable, and obtaining them is itself a research problem, since the labels require private intent that is not recoverable from logs. And its operational facts are derived from the user instruction rather than gold-annotated, so $\sigma_{\mathrm{op}}$ here is a weaker measurement than on \textsc{Bound-Handoff} and its ceiling may be optimistic; $\sigma_b$, the quantity that carries the finding, uses gold markers throughout. Finally, PrivacyLens supplies exactly one norm per datapoint by construction, so it cannot speak to how \emph{frequently} boundary markers appear in real relays, which is a question our setup leaves open.

%% file: tables/ordinal_label_proportions_baseline.tex
\begin{table*}[!ht]
\centering
\small
\setlength{\tabcolsep}{3pt}
\begin{tabular}{@{}llcccccccc@{}}
\toprule
&& \multicolumn{4}{c}{markers ($n=72$)} & \multicolumn{4}{c}{operational ($n=37$)} \\
\cmidrule(lr){3-6}\cmidrule(l){7-10}
\textbf{Model} & \textbf{Condition} & pres & para & weak & abs & pres & para & weak & abs \\
\midrule
GPT-5-mini & \texttt{free\_text} & 0.90 & 0.10 & 0.00 & 0.00 & 1.00 & 0.00 & 0.00 & 0.00 \\
GPT-5-mini & \texttt{compressed\_free\_text} & 0.79 & 0.15 & 0.03 & 0.03 & 0.95 & 0.05 & 0.00 & 0.00 \\
GPT-5-mini & \texttt{preserve\_markers\_instruction} & 0.90 & 0.06 & 0.04 & 0.00 & 0.97 & 0.03 & 0.00 & 0.00 \\
GPT-5-mini & \texttt{sectioned\_template} & 0.96 & 0.03 & 0.01 & 0.00 & 1.00 & 0.00 & 0.00 & 0.00 \\
GPT-5-mini & \texttt{structured\_schema} & 0.79 & 0.19 & 0.00 & 0.01 & 0.95 & 0.05 & 0.00 & 0.00 \\
\addlinespace
DS-R1-32B & \texttt{free\_text} & 0.61 & 0.18 & 0.10 & 0.11 & 0.97 & 0.03 & 0.00 & 0.00 \\
DS-R1-32B & \texttt{compressed\_free\_text} & 0.74 & 0.17 & 0.07 & 0.03 & 0.84 & 0.16 & 0.00 & 0.00 \\
DS-R1-32B & \texttt{preserve\_markers\_instruction} & 0.86 & 0.10 & 0.04 & 0.00 & 0.97 & 0.03 & 0.00 & 0.00 \\
DS-R1-32B & \texttt{sectioned\_template} & 0.82 & 0.17 & 0.01 & 0.00 & 0.97 & 0.00 & 0.03 & 0.00 \\
DS-R1-32B & \texttt{structured\_schema} & 0.53 & 0.39 & 0.04 & 0.04 & 0.86 & 0.11 & 0.03 & 0.00 \\
\addlinespace
Qwen3-32B & \texttt{free\_text} & 0.78 & 0.21 & 0.01 & 0.00 & 1.00 & 0.00 & 0.00 & 0.00 \\
Qwen3-32B & \texttt{compressed\_free\_text} & 0.64 & 0.33 & 0.01 & 0.01 & 0.97 & 0.03 & 0.00 & 0.00 \\
Qwen3-32B & \texttt{preserve\_markers\_instruction} & 0.89 & 0.07 & 0.04 & 0.00 & 0.97 & 0.03 & 0.00 & 0.00 \\
Qwen3-32B & \texttt{sectioned\_template} & 0.76 & 0.24 & 0.00 & 0.00 & 0.92 & 0.08 & 0.00 & 0.00 \\
Qwen3-32B & \texttt{structured\_schema} & 0.56 & 0.38 & 0.03 & 0.04 & 0.86 & 0.11 & 0.00 & 0.03 \\
\bottomrule
\end{tabular}
\caption{E1 baseline ordinal label proportions by condition. Proportions are reported directly from the four-level judge rubric without collapsing labels into binary outcomes.}
\label{tab:ordinal_label_proportions_baseline}
\end{table*}

%% file: tables/cmh_results.tex
\begin{table}[t]
\centering
\scriptsize
\setlength{\tabcolsep}{2pt}
\begin{tabular}{@{}llcccc@{}}
\toprule
\textbf{Model} & \textbf{Test} & \textbf{MH OR [95\% CI]} & $\chi^2$ & \textbf{df} & $p$ \\
\midrule
GPT-5-mini & M vs. O (25w) & 3.92 [1.39, 11.03] & 10.80 & 1 & $0.001$ \\
GPT-5-mini & 25w vs. free (M) & 6.54 [2.40, 17.78] & 20.25 & 1 & $6.8\mathrm{e}{-06}$ \\
DS-R1-32B & M vs. O (25w) & 4.36 [1.68, 11.31] & 14.69 & 1 & $1.3\mathrm{e}{-04}$ \\
DS-R1-32B & 25w vs. free (M) & 6.00 [2.40, 15.01] & 11.07 & 1 & $8.8\mathrm{e}{-04}$ \\
Qwen3-32B & M vs. O (25w) & 4.33 [1.83, 10.25] & 17.82 & 1 & $2.4\mathrm{e}{-05}$ \\
Qwen3-32B & 25w vs. free (M) & 7.22 [3.14, 16.61] & 32.67 & 1 & $1.1\mathrm{e}{-08}$ \\
\bottomrule
\end{tabular}
\caption{Scenario-stratified Cochran--Mantel--Haenszel tests for ordinal-label degradation. Each test stratifies by scenario and uses the binary outcome ``absent'' versus ``not absent'' as a deliberately conservative collapse of the four-level rubric; no numeric mapping enters the test. The marker-vs.-operational rows compare item kinds under 25-word compression, and the marker-compression rows compare 25-word compression against \texttt{free\_text} within marker items.}
\label{tab:cmh_ordinal_absent}
\end{table}